\documentclass[]{template/ovb_arxiv_style}

\usepackage{amsmath}
\usepackage{amssymb}
\usepackage{amsfonts}
\usepackage{graphicx}
\usepackage{booktabs}
\usepackage{array}
\usepackage{multirow}
\usepackage{xcolor}
\usepackage{xspace}
\usepackage{fontawesome5}
\usepackage{pgfplots}
\pgfplotsset{compat=1.17}
\usepgfplotslibrary{groupplots}

\definecolor{armA}{rgb}{0.80,0.33,0.10}
\definecolor{armB}{rgb}{0.03,0.49,0.51}
\definecolor{ovbyellow}{HTML}{F5C63A}
\definecolor{ovbdepth}{HTML}{0B7A75}
\definecolor{ovbsem}{HTML}{2E6BC6}
\definecolor{ovbinst}{HTML}{7B4FB5}
\definecolor{ovbref}{HTML}{D2691E}
\definecolor{ovbgen}{HTML}{6B6B6B}
\definecolor{ovbviolet}{HTML}{8080FF}

\newcommand{\authormark}[1]{\textsuperscript{#1}}
\newcommand{\authorentry}[2]{\mbox{#1\authormark{#2}}}
\newcommand{\authorsep}{\hspace{0.78em}}
\newcommand{\corrmark}{\faEnvelope[regular]}
\makeatletter
\newcommand{\corrnote}[1]{\begingroup\let\thefootnote\relax\renewcommand\@makefntext[1]{\noindent##1}\footnotetext{#1}\endgroup}
\makeatother

\newcommand{\ovblogo}[1]{\includegraphics[height=#1]{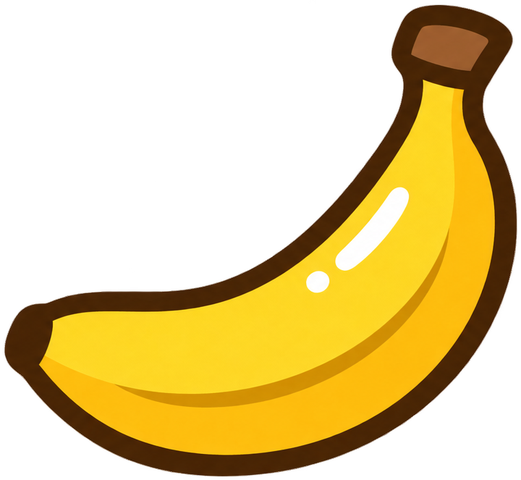}}

\setleftheadercontent{}
\setrightheadericon{}
\setrunningheadericon{\adjustbox{valign=c}{\ovblogo{6mm}}}
\setheadergroupname{Open Vision Banana}

\hypersetup{
  breaklinks=true,
  colorlinks=true,
  citecolor=omnilink,
  linkcolor=omnilink,
  urlcolor=omnilink,
  pdftitle={Open Vision Banana: An Open Framework for Unified Visual Perception through Generation in RGB Space},
  pdfauthor={Xu Lin, Ruilin Yao, Haoran Xu, Zixiang Wang, Ming Ma, Yu Zhao, Pu Miao, Chonghan Liu}
}

\newcommand{\ovbtitlelogo}{\raisebox{-0.22\height}{\includegraphics[height=6.9mm]{figures/logo_banana.png}}\hspace{2.2mm}}
\title{{\fontsize{16}{19}\selectfont\texorpdfstring{\ovbtitlelogo}{}Open Vision Banana: An Open Framework for Unified Visual Perception through Generation in RGB Space}}

\makeatletter
\def\authorlist{%
  {\fontsize{12}{14}\selectfont\color{black!96}\sffamily\bfseries
  \parbox{0.96\linewidth}{\centering
    \makebox[\linewidth][c]{%
      \authorentry{Xu Lin}{1}\authorsep
      \authorentry{Ruilin Yao}{2,9}\authorsep
      \authorentry{Haoran Xu}{3}\authorsep
      \authorentry{Zixiang Wang}{4}%
    }\par\vspace{2.2pt}%
    \makebox[\linewidth][c]{%
      \authorentry{Ming Ma}{5}\authorsep
      \authorentry{Yu Zhao}{6}\authorsep
      \authorentry{Pu Miao}{7}\authorsep
      \authorentry{Chonghan Liu}{8,\corrmark}%
    }%
  }}%
}
\def\affiliationlist{%
  {\fontsize{9.9}{11.9}\selectfont\color{black!70}\sffamily
  \parbox{0.92\linewidth}{\centering
    \mbox{\authormark{1}Renmin University of China}\authorsep
    \mbox{\authormark{2}Wuhan University of Technology}\authorsep
    \mbox{\authormark{3}Zhejiang University}\authorsep
    \mbox{\authormark{4}China University of Mining and Technology}\authorsep
    \mbox{\authormark{5}Institute of Neuroscience, Chinese Academy of Sciences}\authorsep
    \mbox{\authormark{6}ByteDance}\authorsep
    \mbox{\authormark{7}Meituan}\authorsep
    \mbox{\authormark{8}University of California, Los Angeles}\authorsep
    \mbox{\authormark{9}Institute of Automation, Chinese Academy of Sciences}\par
  }}%
}
\makeatother

\abstract{
Vision Banana showed that a closed image generator, instruction-tuned to paint perception outputs as RGB images, becomes a generalist vision model. Does this carry over to \emph{open} generators, and if not, what fails? We answer with \textbf{Open Vision Banana (OVB)}, an open framework that adapts the 7B diffusion generator of Qwen-Image-2.1 into one instruction-driven model for metric depth, semantic, instance, and referring segmentation, using a single shared low-rank adapter and public data. Each task is a \emph{task contract} whose RGB encoder, instruction, and decoder share one specification and are checked before training. Studying 42 externally evaluated checkpoints from two 100k-update runs, six exact weight-space merges, and eight controlled shorter runs, we find that the gap to the closed system is specific and measurable. Geometry is learned early while absolute scale is not: median-aligned NYUv2 $\delta_1$ stays at 0.89--0.93 throughout training as unaligned $\delta_1$ wanders between 0.11 and 0.50. The scale is not committed by a single sample: a ten-seed ensemble lifts unaligned $\delta_1$ from 0.348 to 0.516 without retraining. Depth and 2D segmentation compete for the shared adapter, and weight merging moves along this trade-off without escaping it. Finally, lower validation flow loss comes with lower Cityscapes accuracy (Spearman $\rho\approx-0.9$ between loss reduction and mIoU), so the loss cannot select checkpoints. We release code, contracts, training data, and all evaluation outputs.
}

\checkdata[Code]{\href{https://github.com/isLinXu/open-vision-banana}{\nolinkurl{github.com/isLinXu/open-vision-banana}}}
\checkdata[Project Page]{\href{https://islinxu.github.io/open-vision-banana/}{\nolinkurl{islinxu.github.io/open-vision-banana}}}
\checkdata[Models]{\href{https://huggingface.co/collections/gatilin/open-vision-banana-model-zoo}{\nolinkurl{huggingface.co/collections/gatilin/open-vision-banana-model-zoo}}}

\begin{document}

\maketitle
\corrnote{\corrmark\ Corresponding author.}

\section{Introduction}
\label{sec:intro}

Can one image generator, guided only by a task instruction, see depth, classes, instances, and referred objects? Generative pretraining makes this plausible: to synthesize realistic scenes, a generator must model appearance, object structure, and spatial layout. Marigold~\cite{ke2024repurposing} showed that a diffusion prior transfers to monocular depth, and Vision Banana~\cite{gabeur2026image} went much further: it paints segmentation, depth, and normals as RGB images, instruction-tunes an image generator on them, and reports results on par with specialist models while preserving image synthesis. Image generation may thus become a general interface for perception.

The evidence for this paradigm, however, rests on a closed generator (Nano Banana Pro) trained on in-house data, with neither weights nor evaluation tools released. Two questions are therefore open: \emph{does the paradigm carry over to open generators}, and, if it falls short, \emph{where exactly does it fail, and why?} We argue that the second question is the more useful one. A reproduction that matches a number tells the community little; a measured account of the failure modes tells the next model what to fix.

\begin{figure*}[t]
\centering
\includegraphics[width=\linewidth]{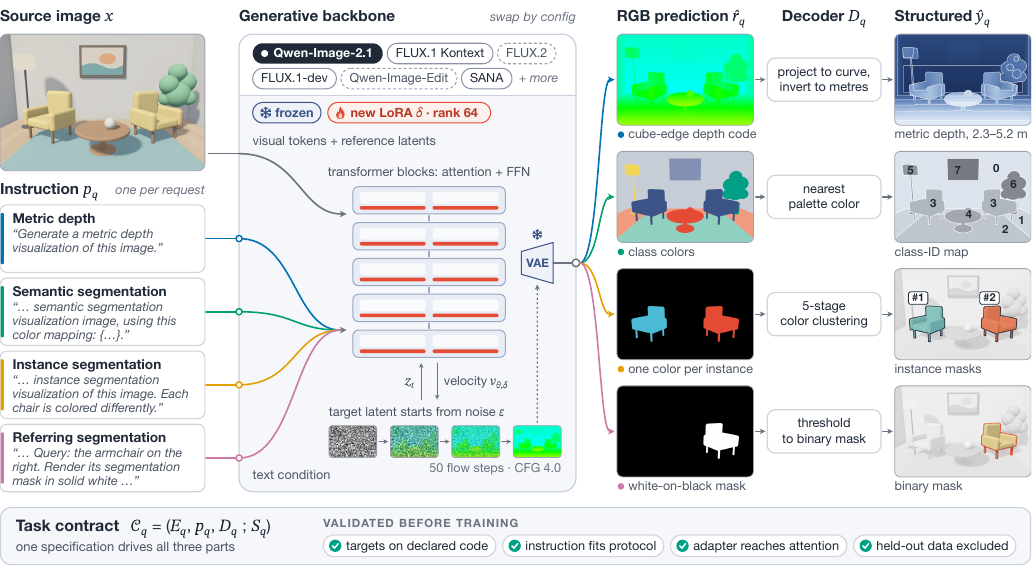}
\caption{\textbf{Open Vision Banana overview.} A source image $x$ and a task instruction $p_q$ condition a frozen generative backbone (Qwen-Image-2.1 in all experiments; the backbone is a configuration choice) that carries one newly initialized rank-64 LoRA adapter $\delta$ shared by all tasks. The source enters as visual tokens and reference latents, the instruction as the text condition, and the target latent is generated from noise in 50 flow steps. The generated RGB image $\hat{r}_q$ is converted by a parameter-free decoder $D_q$ into a structured prediction $\hat{y}_q$: metric depth, a class-ID map, instance masks, or a binary mask. Bottom: a task contract $\mathcal{C}_q=(E_q,p_q,D_q;S_q)$ ties encoder, instruction, and decoder to one specification and is validated before training. The scene is a rendered illustration; its RGB tiles are produced by the contract encoders from exact annotations and are not model predictions.}
\label{fig:framework}
\end{figure*}

We address both with \textbf{Open Vision Banana (OVB)} (Figure~\ref{fig:framework}), an open, independent framework that realizes the paradigm on Qwen-Image-2.1~\cite{wu2025qwenimage,zhao2026qwenimage2}---a unified generation-and-editing model whose 7B diffusion transformer is conditioned by a frozen 8B vision-language encoder---using public data. Building it first exposes an engineering obstacle that silently breaks generative perception: RGB is only a carrier, and a prediction is meaningful only if the target encoder, the instruction, and the decoder agree on units, color codes, and mask conventions. When they disagree, the training loss still decreases while the decoded predictions fail as a whole. OVB therefore writes every task as a \textbf{task contract} that derives encoder, instruction, and decoder from one specification and checks them before training. A frozen editing backbone with one shared low-rank adapter then serves all tasks, selected only by the instruction.

The open setting lets us study the paradigm at a granularity that a closed system does not allow: two 100k-update runs that differ only in learning rate, with 42 checkpoints scored on six benchmarks by an external suite; six exact weight-space merges of these checkpoints; and eight shorter runs that vary the starting point, the task mixture, the depth colormap, and the adaptation mode. OVB stays below the closed reference on every benchmark (Table~\ref{tab:main}), but the shortfall is neither uniform nor mysterious. It decomposes into five findings:

\begin{itemize}
  \item \textbf{Geometry is learned, scale is not.} Median-aligned depth is near the reference from early on; almost all of the depth gap is one number per image (Sec.~\ref{sec:f-scale}).
  \item \textbf{The scale is not committed.} The same input yields different scales under different noise; a seed ensemble recovers much of the gap without retraining (Sec.~\ref{sec:f-variance}).
  \item \textbf{Tasks compete for one adapter.} Depth improves while Cityscapes and ReasonSeg decline, along training and across mixtures and learning rates; full fine-tuning erases referring ability (Sec.~\ref{sec:f-tradeoff}).
  \item \textbf{Merging does not escape the trade-off.} Exact merges move along it; only semantic segmentation gains beyond interpolation (Sec.~\ref{sec:f-merge}).
  \item \textbf{The training loss is blind to what fails.} Validation flow loss falls while Cityscapes accuracy falls with it (Sec.~\ref{sec:f-loss}).
\end{itemize}

\noindent In summary, we contribute:
\begin{enumerate}
  \item \textbf{An open framework for generation-as-perception}: the Vision Banana paradigm on a public editing backbone with one shared adapter for four tasks, made reliable by task contracts that are checked before training.
  \item \textbf{A diagnostic study} that locates the gap to the closed system in absolute depth scale, sampling variance, and task competition under a shared adapter, with every claim backed by full-benchmark external evaluation.
  \item \textbf{A reproducible baseline}: code, contracts, the 295{,}505-row referring-segmentation training pool, leakage-audited evaluation protocols, and every evaluation output with checksums, with all deviations from the reference protocol documented.
\end{enumerate}

\section{Related Work}
\label{sec:related}

\paragraph{Generalist vision interfaces.}
Painter~\cite{wang2023painter} casts dense prediction as image inpainting with in-context examples. InstructCV~\cite{gan2024instructcv} and InstructDiffusion~\cite{geng2024instructdiffusion} instruction-tune text-to-image diffusion models to emit task outputs as images, building on the editing interface of InstructPix2Pix~\cite{brooks2023instructpix2pix}. Vision Banana~\cite{gabeur2026image} scales this idea to a strong closed generator and reports specialist-level results, and related studies find emergent perception in image and video generators~\cite{wiedemer2025video,zuo2025nano}. OVB follows the RGB formulation of Vision Banana on public models and data; everything the paper does not disclose (adapter, optimizer, schedule, mixture, resolution, sampling) is our choice. Unlike prior generalists, we target \emph{metric} depth and analyze why the open recipe falls short rather than proposing a new architecture.

\paragraph{Generative priors for dense prediction.}
Marigold~\cite{ke2024repurposing}, GeoWizard~\cite{fu2024geowizard}, GenPercept~\cite{xu2025genpercept}, StableNormal~\cite{ye2024stablenormal}, and Lotus~\cite{he2025lotus} repurpose diffusion models for depth and normals, mostly predicting relative or affine-invariant quantities with a task-specific model. StableNormal identifies sampling variance as an obstacle for diffusion-based geometry; we find that in the metric setting this variance concentrates in absolute scale. Specialist metric-depth models reach high zero-shot accuracy with dedicated architectures, large mixed datasets, and often camera intrinsics~\cite{bochkovskiy2025depth,piccinelli2025unik3d,wang2026moge,lin2025depth,yeh2026depthlm}; OVB uses no intrinsics and no depth head.

\paragraph{Referring segmentation.}
Recent systems couple a language model with a mask decoder~\cite{lai2024lisa,wang2026x,wei2024hyperseg} or a promptable segmenter~\cite{carion2026sam}; OVB paints the mask as an image with the same adapter that paints depth.

\paragraph{Adaptation and weight averaging.}
OVB trains a rank-stabilized LoRA~\cite{hu2021lora,kalajdzievski2023rank} with conditional flow matching~\cite{lipman2022flow}. Averaging fine-tuned weights that share an initialization can combine their abilities~\cite{wortsman2022model}; we test whether it resolves the task trade-off of a shared adapter.

\paragraph{What this paper measures.}
Each failure we report has precedent---hard metric scale, sampling variance in diffusion geometry~\cite{ye2024stablenormal}, capacity trade-offs---so what we establish is their size relative to one another inside one open system, over 42 externally scored checkpoints.

\section{Open Vision Banana}
\label{sec:method}

\subsection{Task Contracts}

\begin{figure*}[t]
\centering
\includegraphics[width=\linewidth]{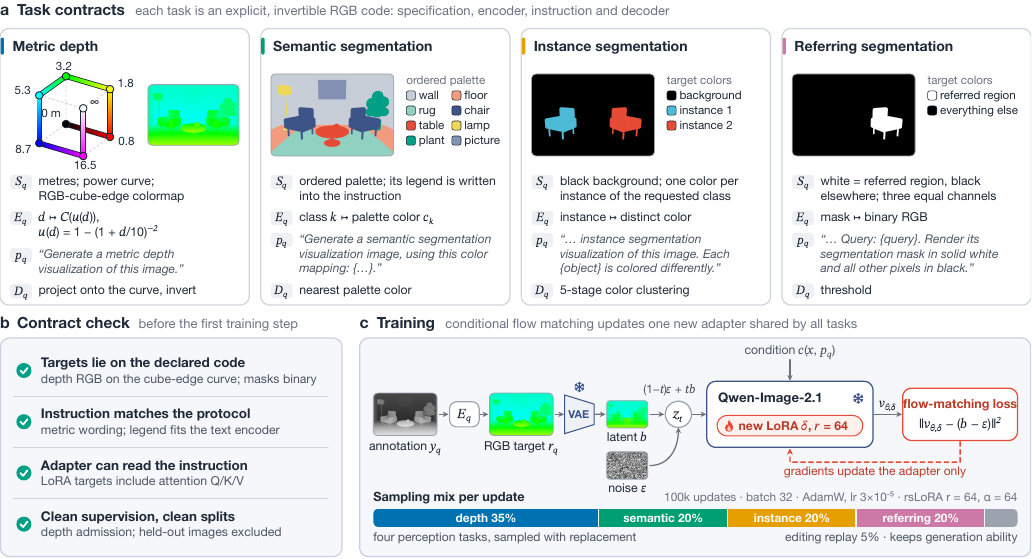}
\caption{\textbf{Task contracts, contract check, and training.} (a) Each task is an explicit, decodable RGB code given by a specification $S_q$, an encoder $E_q$, an instruction $p_q$, and a decoder $D_q$; the depth card shows the colormap path along the edges of the RGB cube, each vertex labeled with the metric depth it encodes under Eq.~\eqref{eq:depth-transform}. (b) Checks that run on the training manifest and configuration before the first update. (c) Training: an annotation is encoded into an RGB target and then into a latent $b$ by the frozen autoencoder; conditional flow matching updates the shared adapter while the backbone stays frozen. Run~A uses peak lr $3\times10^{-5}$ (shown) and run~B uses $1\times10^{-5}$; editing replay is included with the intent of preserving generation, which we have not measured (Sec.~\ref{sec:discussion}).}
\label{fig:contracts}
\end{figure*}

OVB returns every prediction as an RGB image. Given an image $x$ and a task $q$,
\begin{equation}
  \hat{r}_q = G_{\theta,\delta}(x,\, p_q), \qquad \hat{y}_q = D_q(\hat{r}_q),
  \label{eq:interface}
\end{equation}
where $\theta$ are the frozen backbone parameters, $\delta$ is the adapter shared by all tasks, $p_q$ is the instruction, and $D_q$ is the task decoder. In training, the annotation $y_q$ becomes the RGB target $r_q = E_q(y_q)$. A prediction is meaningful only if $E_q$, $p_q$, and $D_q$ agree, so OVB derives all three from one specification $S_q$ and calls $\mathcal{C}_q = (E_q, p_q, D_q; S_q)$ the \textbf{task contract} of task $q$. $S_q$ records every convention: the unit, curve, and colormap of depth; the ordered class--color palette; the foreground and background colors of masks. Tasks differ only in their contracts (Table~\ref{tab:contracts}, Figure~\ref{fig:contracts}).

\begin{table*}[t]
\centering
\small
\caption{\textbf{Task contracts.} Encoding and decoding share the specification $S_q$; the instruction templates are used unchanged in training and evaluation (shortened in Figures~\ref{fig:framework}--\ref{fig:contracts} for space; all templates are listed in \cref{sec:s-prompts}).}
\label{tab:contracts}
\setlength{\tabcolsep}{3pt}
\begin{tabular}{l>{\raggedright\arraybackslash}p{2.8cm}>{\raggedright\arraybackslash}p{2cm}>{\raggedright\arraybackslash}p{6.5cm}>{\raggedright\arraybackslash}p{2.5cm}}
\toprule
Task & Specification $S_q$ & Encoding $E_q$ & Instruction template $p_q$ & Decoding $D_q$ \\
\midrule
Depth & Metric units; power transform; cube-edge colormap & $d\mapsto C(u(d))$ & \textit{Generate a metric depth visualization of this image using the cube\_edge colormap.} & Project to curve, invert \\
Semantic & Ordered palette $\{c_k\}_{k=0}^{K-1}$ & Class $k\mapsto c_k$ & \textit{Generate a semantic segmentation visualization image, using this color mapping: \{road: ..., ...\}.} & Nearest palette color \\
Instance & Black bg.; one color per instance & Instance $\mapsto$ distinct color & \textit{Generate an instance segmentation visualization of this image. Each \{object\} is colored differently.} & 5-stage color clustering \\
Referring & White fg., black bg. & Mask $\mapsto$ binary RGB & \textit{Identify the object or region referred to by the following query in the image. Query: \{query\} Render its segmentation mask in solid white and all other pixels in black.} & Threshold \\
\bottomrule
\end{tabular}
\end{table*}

\paragraph{Decodable RGB codes.}
Each task uses a deterministic decoder matched to its RGB encoding. For metric depth $d$ in meters, the encoder first applies
\begin{equation}
  u(d) = 1 - \Bigl(1 + \tfrac{d}{10}\Bigr)^{-2},
  \label{eq:depth-transform}
\end{equation}
which maps $[0,\infty)$ to $[0,1)$, with half of the code range allocated to depths below $10(\sqrt{2}-1)\approx 4.1\,\mathrm{m}$. It then maps $u$ to a piecewise-linear path $C$ along seven edges of the RGB cube, providing $7\times255=1{,}785$ 8-bit steps. At inference, each output pixel is projected onto $C$ to recover $\hat{u}$, and metric depth is reconstructed by
\[
  \hat{d}=10\bigl((1-\hat{u})^{-1/2}-1\bigr).
\]
Thus, changes in absolute depth scale alter the encoded colors through the nonlinear depth transform. For semantic segmentation, each pixel is assigned the class ID of its nearest palette color under Euclidean distance in RGB space. For instance segmentation, the decoder proceeds through five stages: background initialization, 16-connected seed-based color grouping, noise pruning, boundary-artifact removal, and spatially constrained merging; its stages and thresholds are listed in \cref{sec:s-instance}. For referring segmentation, pixels whose mean RGB value exceeds $0.5$ are classified as foreground, and all connected components are retained.

\paragraph{Validation before training.}
Before the first update, the framework checks the manifest and configuration against the contracts (Figure~\ref{fig:contracts}(b)): (i)~targets lie on the declared code (depth targets survive a decode--re-encode round trip within $2/255$, referring targets are binary, storage is lossless); (ii)~instructions match the protocol (a relative-depth instruction is rejected under metric decoding, and no color legend exceeds the backbone's text-encoder length); (iii)~the adapter covers the attention projections that read the instruction, since an adapter restricted to feed-forward layers can only memorize fixed colors; and (iv)~supervision and splits are clean (depth frames with too few valid pixels or large encoding error are rejected, and training images that overlap an evaluation set are removed by image identifier). Each check targets a failure that otherwise appears only after training.

\subsection{Backbone and Shared Adaptation}

\paragraph{Conditioning.}
OVB builds on the editing model Qwen-Image-2.1, which accepts a source image natively: $x$ enters as visual tokens and reference latents, $p_q$ is the text condition, and the target is generated from noise under $c(x,p_q)$, the interface the backbone already uses for editing. Training uses conditional flow matching,
\begin{equation}
  \mathcal{L} = \mathbb{E}_{t,\epsilon}\,\bigl\|v_{\theta,\delta}(z_t,t,c(x,p_q)) - (b-\epsilon)\bigr\|_2^2,\quad z_t=(1-t)\epsilon+t\,b,
  \label{eq:flow}
\end{equation}
where $b$ is the autoencoder latent of the RGB target and $t=\mathrm{sigmoid}(\eta)$, $\eta\sim\mathcal{N}(0,1)$. All tasks, replay included, use this loss with unit weight and no auxiliary term.

\paragraph{Shared adapter and starting point.}
The transformer, autoencoder, and text encoder are frozen. The frozen transformer is a fused checkpoint: two adapters from earlier OVB stages, trained from the same parent checkpoint (the parent of Table~\ref{tab:family}), are merged into it with weights 0.6 and 0.4: the depth-heavy run C1 of Table~\ref{tab:family} and a rank-128 adapter trained for 20k updates on a uniform mixture of depth, normals, semantic, instance, and referring segmentation (18\% each) with 10\% editing replay. On top we train one zero-initialized rank-stabilized LoRA ($r=64$, $\alpha=64$, scale $\alpha/\sqrt{r}=8$) on all attention projections and feed-forward layers; it is the only trained parameter set, and switching tasks changes only the instruction.

\paragraph{Optimization.}
Both main runs train for 100{,}000 updates with AdamW~\cite{loshchilov2019decoupled} (weight decay 0.01), global batch 32 on 16 GPUs, resolution 512, 2{,}000 warmup updates, cosine decay, gradient clipping at norm 1, conditioning dropout 0.05, and an adapter EMA with decay 0.9999. They differ only in the peak learning rate: $3\times10^{-5}$ (run~A) and $1\times10^{-5}$ (run~B). The validation flow loss is evaluated every 2{,}500 updates; ``best'' denotes the EMA adapter saved when this loss reaches a new minimum. All configurations are released.

\subsection{Training Data and Target Construction}
\label{sec:data-construction}

\begin{figure*}[t]
\centering
\adjustbox{max width=\linewidth}{
\begin{tikzpicture}[font=\footnotesize]
\node[anchor=west, font=\small\bfseries] at (-2.6, 0.62) {(a) Training pool by task and source (records)};
\node[anchor=east] at (-0.12, 0.000) {\textbf{Metric depth}};
\fill[ovbdepth!100] (0.000, -0.250) rectangle (6.169, 0.250);
\draw[white, line width=0.6pt] (6.169, -0.250) -- (6.169, 0.250);
\node[text=white, font=\scriptsize, align=center, inner sep=0pt] at (3.085, 0.000) {Indoor observed\\[-2.5pt]{\tiny 292.2k}};
\fill[ovbdepth!78] (6.169, -0.250) rectangle (7.225, 0.250);
\draw[white, line width=0.6pt] (7.225, -0.250) -- (7.225, 0.250);
\fill[ovbdepth!60] (7.225, -0.250) rectangle (8.041, 0.250);
\draw[white, line width=0.6pt] (8.041, -0.250) -- (8.041, 0.250);
\fill[ovbdepth!45] (8.041, -0.250) rectangle (8.487, 0.250);
\draw[white, line width=0.6pt] (8.487, -0.250) -- (8.487, 0.250);
\fill[ovbdepth!32] (8.487, -0.250) rectangle (8.600, 0.250);
\draw[white, line width=0.6pt] (8.600, -0.250) -- (8.600, 0.250);
\node[anchor=west, font=\footnotesize] at (8.660, 0.000) {\textbf{407{,}278}};
\node[anchor=west, text width=3.9cm, align=left, font=\scriptsize, text=black!60, inner sep=1pt] at (10.200, 0.000) {LingBot-Depth 50.0k, TartanAir 38.7k, Hypersim 21.1k, DIODE (added) 5.3k};
\node[anchor=east] at (-0.12, -0.780) {\textbf{Semantic}};
\fill[ovbsem!100] (0.000, -1.030) rectangle (2.257, -0.530);
\draw[white, line width=0.6pt] (2.257, -1.030) -- (2.257, -0.530);
\node[text=white, font=\scriptsize, align=center, inner sep=0pt] at (1.129, -0.780) {COCONut\\[-2.5pt]{\tiny 106.9k}};
\fill[ovbsem!78] (2.257, -1.030) rectangle (3.083, -0.530);
\draw[white, line width=0.6pt] (3.083, -1.030) -- (3.083, -0.530);
\node[anchor=west, font=\footnotesize] at (3.143, -0.780) {\textbf{146{,}027}};
\node[anchor=west, text width=3.9cm, align=left, font=\scriptsize, text=black!60, inner sep=1pt] at (10.200, -0.780) {Earlier pool 39.1k};
\node[anchor=east] at (-0.12, -1.560) {\textbf{Instance}};
\fill[ovbinst!100] (0.000, -1.810) rectangle (4.108, -1.310);
\draw[white, line width=0.6pt] (4.108, -1.810) -- (4.108, -1.310);
\node[text=white, font=\scriptsize, align=center, inner sep=0pt] at (2.054, -1.560) {COCONut\\[-2.5pt]{\tiny 194.5k}};
\fill[ovbinst!78] (4.108, -1.810) rectangle (4.923, -1.310);
\draw[white, line width=0.6pt] (4.923, -1.810) -- (4.923, -1.310);
\node[anchor=west, font=\footnotesize] at (4.983, -1.560) {\textbf{233{,}133}};
\node[anchor=west, text width=3.9cm, align=left, font=\scriptsize, text=black!60, inner sep=1pt] at (10.200, -1.560) {Earlier pool 38.6k};
\node[anchor=east] at (-0.12, -2.340) {\textbf{Referring}};
\fill[ovbref!100] (0.000, -2.590) rectangle (2.169, -2.090);
\draw[white, line width=0.6pt] (2.169, -2.590) -- (2.169, -2.090);
\node[text=white, font=\scriptsize, align=center, inner sep=0pt] at (1.084, -2.340) {RefCOCO\\[-2.5pt]{\tiny 102.7k}};
\fill[ovbref!78] (2.169, -2.590) rectangle (4.185, -2.090);
\draw[white, line width=0.6pt] (4.185, -2.590) -- (4.185, -2.090);
\node[text=white, font=\scriptsize, align=center, inner sep=0pt] at (3.177, -2.340) {RefCOCO+\\[-2.5pt]{\tiny 95.5k}};
\fill[ovbref!60] (4.185, -2.590) rectangle (5.705, -2.090);
\draw[white, line width=0.6pt] (5.705, -2.590) -- (5.705, -2.090);
\node[text=white, font=\scriptsize, align=center, inner sep=0pt] at (4.945, -2.340) {RefCOCOg\\[-2.5pt]{\tiny 72.0k}};
\fill[ovbref!45] (5.705, -2.590) rectangle (5.990, -2.090);
\draw[white, line width=0.6pt] (5.990, -2.590) -- (5.990, -2.090);
\fill[ovbref!32] (5.990, -2.590) rectangle (6.240, -2.090);
\draw[white, line width=0.6pt] (6.240, -2.590) -- (6.240, -2.090);
\node[anchor=west, font=\footnotesize] at (6.300, -2.340) {\textbf{295{,}505}};
\node[anchor=west, text width=3.9cm, align=left, font=\scriptsize, text=black!60, inner sep=1pt] at (10.200, -2.340) {LLM-Seg 13.5k, Rea2Seg 11.8k};
\node[anchor=east] at (-0.12, -3.120) {\textbf{Editing replay}};
\fill[ovbgen!100] (0.000, -3.370) rectangle (0.348, -2.870);
\draw[white, line width=0.6pt] (0.348, -3.370) -- (0.348, -2.870);
\node[anchor=west, font=\footnotesize] at (0.408, -3.120) {\textbf{16{,}492}};
\node[anchor=west, text width=3.9cm, align=left, font=\scriptsize, text=black!60, inner sep=1pt] at (10.200, -3.120) {Editing pairs 16.5k};
\node[anchor=west, font=\small\bfseries] at (-2.6, -4.100) {(b) Share of the pool vs.\ sampling probability per update};
\node[anchor=west, font=\scriptsize, align=left, text=black!70] at (9.850, -4.100) {expected passes per record};
\node[anchor=east] at (-0.12, -4.650) {\textbf{Metric depth}};
\fill[ovbdepth!35] (0, -4.630) rectangle (7.972, -4.430);
\fill[ovbdepth] (0, -4.870) rectangle (7.525, -4.670);
\node[anchor=west, font=\scriptsize, text=black!60] at (8.022, -4.530) {37.1\%};
\node[anchor=west, font=\scriptsize] at (7.575, -4.770) {\textbf{35\%}};
\node[anchor=west, font=\footnotesize] at (9.850, -4.650) {$\times$2.75};
\node[anchor=east] at (-0.12, -5.430) {\textbf{Semantic}};
\fill[ovbsem!35] (0, -5.410) rectangle (2.858, -5.210);
\fill[ovbsem] (0, -5.650) rectangle (4.300, -5.450);
\node[anchor=west, font=\scriptsize, text=black!60] at (2.908, -5.310) {13.3\%};
\node[anchor=west, font=\scriptsize] at (4.350, -5.550) {\textbf{20\%}};
\node[anchor=west, font=\footnotesize] at (9.850, -5.430) {$\times$4.38};
\node[anchor=east] at (-0.12, -6.210) {\textbf{Instance}};
\fill[ovbinst!35] (0, -6.190) rectangle (4.563, -5.990);
\fill[ovbinst] (0, -6.430) rectangle (4.300, -6.230);
\node[anchor=west, font=\scriptsize, text=black!60] at (4.613, -6.090) {21.2\%};
\node[anchor=west, font=\scriptsize] at (4.350, -6.330) {\textbf{20\%}};
\node[anchor=west, font=\footnotesize] at (9.850, -6.210) {$\times$2.75};
\node[anchor=east] at (-0.12, -6.990) {\textbf{Referring}};
\fill[ovbref!35] (0, -6.970) rectangle (5.784, -6.770);
\fill[ovbref] (0, -7.210) rectangle (4.300, -7.010);
\node[anchor=west, font=\scriptsize, text=black!60] at (5.834, -6.870) {26.9\%};
\node[anchor=west, font=\scriptsize] at (4.350, -7.110) {\textbf{20\%}};
\node[anchor=west, font=\footnotesize] at (9.850, -6.990) {$\times$2.17};
\node[anchor=east] at (-0.12, -7.770) {\textbf{Editing replay}};
\fill[ovbgen!35] (0, -7.750) rectangle (0.323, -7.550);
\fill[ovbgen] (0, -7.990) rectangle (1.075, -7.790);
\node[anchor=west, font=\scriptsize, text=black!60] at (0.373, -7.650) {1.5\%};
\node[anchor=west, font=\scriptsize] at (1.125, -7.890) {\textbf{5\%}};
\node[anchor=west, font=\footnotesize] at (9.850, -7.770) {$\times$9.70};
\fill[black!30] (0, -8.380) rectangle (0.3, -8.220);
\node[anchor=west, text=black!65, font=\scriptsize] at (0.35, -8.300) {share of the 1{,}098{,}435 records};
\fill[black!80] (4.75, -8.380) rectangle (4.5, -8.220);
\node[anchor=west, text=black!65, font=\scriptsize] at (5.1, -8.300) {sampling probability (100k updates $\times$ batch 32 = 3.2M draws)};
\end{tikzpicture}}
\caption{\textbf{Training data and task mixture of the main runs.} (a)~The 1{,}098{,}435 training records by task and source; one record is an image--instruction--target triple, so an image can appear in several records. Small sources are listed to the right of their bar; ``Indoor observed'' combines ARKitScenes, DIODE, and SUN RGB-D frames with at least 98\% observed depth, and ``Earlier pool'' is the segmentation data of the earlier OVB stages. (b)~Each task's share of the pool (light) against its per-update sampling probability (dark), and the expected number of times each record is drawn over the 3.2M draws of a 100k-update run. Sampling reweights the pool: semantic segmentation and editing replay are drawn more often than their size implies, referring segmentation less often.}
\label{fig:data}
\end{figure*}
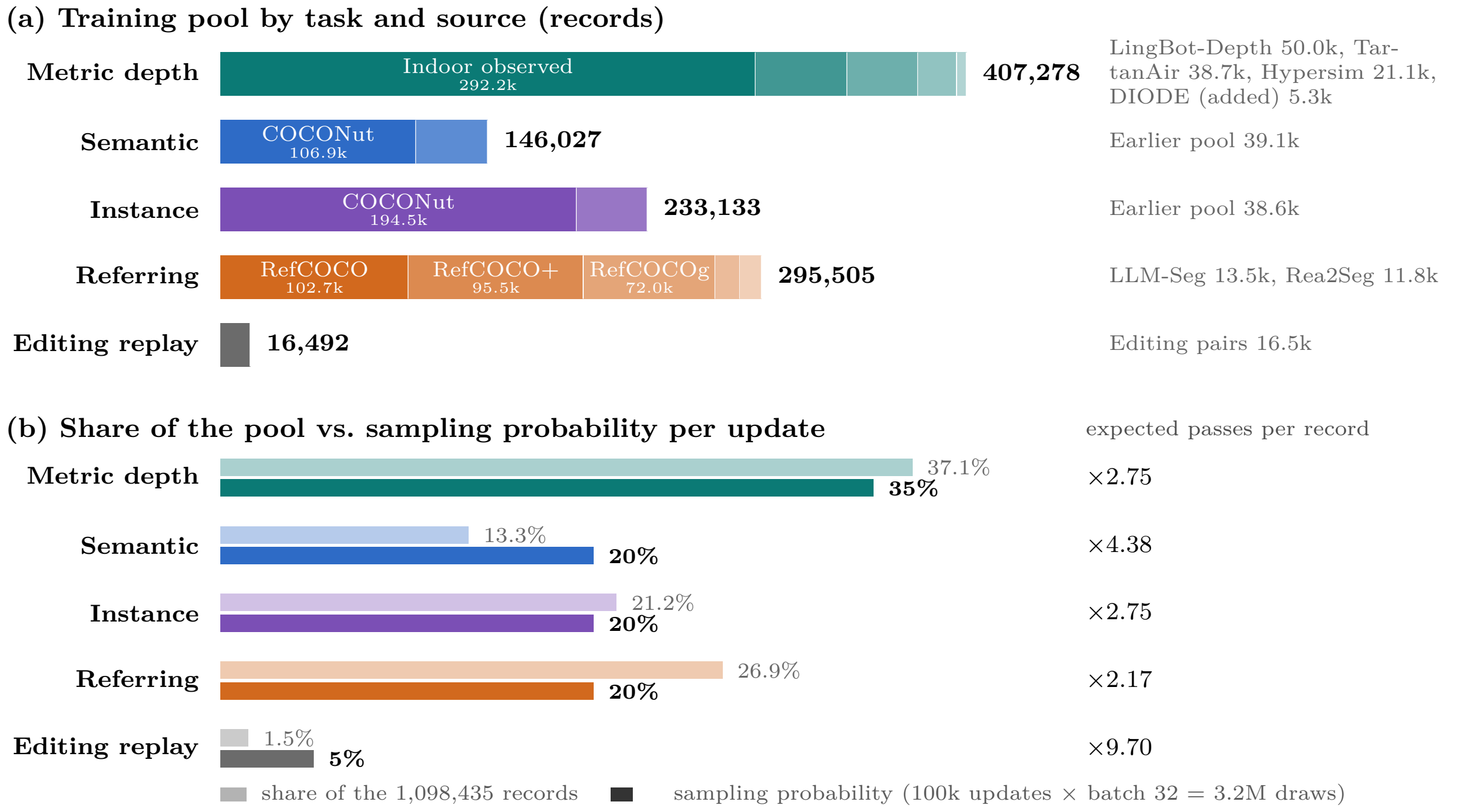

The training pool contains 1{,}098{,}435 image--instruction--target records (\cref{fig:data}). Multiple tasks or queries may share a source image, so these counts describe supervision pairs. Each update samples a task with probability 0.35 for metric depth, 0.20 each for semantic, instance, and referring segmentation, and 0.05 for image-editing replay; records are sampled uniformly within each task. The replay pool contains 16{,}492 editing pairs and is included with the intent of preserving generation ability.

The 407{,}278 depth records combine TartanAir \cite{wang2020tartanair}, Hypersim \cite{roberts2021hypersim}, LingBot-Depth \cite{tan2026masked}, ARKitScenes \cite{baruch2021arkitscenes}, SUN RGB-D (excluding NYU-derived samples) \cite{song2015sun}, and DIODE \cite{vasiljevic2019diode} data. Native-depth admission checks cover 359{,}693 records, while 47{,}585 retained legacy records are explicitly exempt from these checks. For newly prepared indoor and DIODE targets, invalid depth pixels are encoded as black; validity masks support data admission but do not mask the training loss.

The segmentation pools contain 146{,}027 semantic and 233{,}133 instance records, including category-conditioned targets reconstructed from COCONut \cite{deng2024coconut} annotations. In this component, each image contributes one semantic view containing at most 12 eligible classes and up to three category-specific instance views; unselected regions are treated as background relative to the requested categories. Semantic legends use one of four formats, and colors are remapped with probability 0.5 to encourage reliance on the instruction-specified color mapping. Targets are stored as lossless RGB images consistent with the task contracts.

The 295{,}505 referring records comprise 270{,}195 RefCOCO-family pairs \cite{yu2016modeling}, 11{,}812 Rea2Seg pairs, and 13{,}498 LLM-Seg COCO pairs. Rea2Seg \cite{gao2026reason} contributes its GQA, LISA+, and spatial subsets, excluding its ReasonSeg category. LLM-Seg \cite{wang2024llm} contributes released questions paired with native RLE masks, without answer or chain-of-thought conditioning. Its retained subset is selected through held-out image exclusion, native mask validity and foreground-area checks, and visibility filtering under the 512-pixel training crop. Source-provided masks are converted into the white-foreground, black-background targets specified by the referring task contract.

\subsection{Inference and Evaluation Protocol}
\label{sec:protocol}

Inference generates the target from noise with 50 sampling steps, true classifier-free guidance~\cite{ho2022classifier} 4.0 against an empty negative prompt, seed 42, and a $512\times512$ canvas; predictions are decoded and then resized to the native ground-truth resolution (nearest neighbor for labels, bilinear for depth). Metric depth is scored on NYUv2~\cite{silberman2012indoor} (654 images, Eigen crop~\cite{eigen2014depth}, 10\,m cap), ETH3D~\cite{schops2017multi} (454, 80\,m cap), and DIODE-Indoor~\cite{vasiljevic2019diode} (325, no cap) with $\delta_1$ and AbsRel accumulated over pixels and \emph{without any alignment}; we additionally report a per-image median-aligned $\delta_1$ as a diagnostic that uses the ground truth; it corrects a scale but not a shift, unlike the affine-invariant protocol of~\cite{ke2024repurposing}, and is not comparable to the reference. Cityscapes val~\cite{cordts2016cityscapes} (500) is scored with mIoU over 19 classes from a global confusion matrix, RefCOCOg~\cite{mao2016generation} UMD val~\cite{nagaraja2016modeling} (4{,}896 expressions) with cIoU, and ReasonSeg val~\cite{lai2024lisa} (344 queries, original queries, \texttt{legacy\_union} masks) with gIoU. Training and evaluation images are disjoint by image identifier; NYUv2, ETH3D, and ReasonSeg contribute no training data at all, while DIODE and RefCOCOg contribute their training splits and are reported as same-source held-out results. All scores come from the same evaluation protocol; per-benchmark details are in \cref{sec:s-protocol}.

\paragraph{Comparability with the reference.}
Depth, Cityscapes, and RefCOCOg follow the reference definitions, but the reference does not state crops, caps, aggregation, resolution, or sampling settings, so these are our choices. Two benchmarks are \emph{not} comparable: Vision Banana rewrites ReasonSeg queries with Gemini~2.5 Pro before segmentation~\cite{gabeur2026image}, and we pass the original query. We report the reference numbers as context, not as a matched comparison.

\section{Main Results}
\label{sec:main}

\begin{table*}[t]
\centering
\small
\caption{\textbf{OVB endpoints on the trade-off, with the closed reference as context.} Depth is unaligned $\delta_1$ (single sample); per-image median-aligned $\delta_1$, a ground-truth diagnostic, is in parentheses. ``Start'' is the fused base before the shared adapter is trained. Balanced: run~B, 17.5k-update EMA adapter; Depth: run~B, final 100k-update adapter. Bold marks the best same-protocol OVB score per column. Vision Banana numbers are copied from~\cite{gabeur2026image}. $^\ast$Uses a query-rewriting language model; not comparable (Sec.~\ref{sec:protocol}).}
\label{tab:main}
\setlength{\tabcolsep}{4.5pt}
\adjustbox{max width=\linewidth}{%
\begin{tabular}{lcccccc}
\toprule
 & \multicolumn{3}{c}{Metric depth $\delta_1$ (aligned)} & Semantic & \multicolumn{2}{c}{Referring} \\
\cmidrule(lr){2-4}\cmidrule(lr){5-5}\cmidrule(lr){6-7}
Model & NYUv2 & ETH3D & DIODE-In. & Cityscapes mIoU & RefCOCOg cIoU & ReasonSeg gIoU \\
\midrule
OVB start (fused base) & 0.278 (0.873) & \textbf{0.292} (0.924) & 0.300 (0.844) & \textbf{0.386} & 0.466 & 0.417 \\
OVB-Balanced (B, 17.5k EMA) & 0.407 (0.921) & 0.176 (0.932) & 0.277 (0.896) & 0.355 & \textbf{0.518} & \textbf{0.431} \\
OVB-Depth (B, 100k final) & \textbf{0.497} (0.926) & 0.225 (0.974) & \textbf{0.353} (0.926) & 0.298 & 0.481 & 0.333 \\
\midrule
\textit{Vision Banana (closed)}~\cite{gabeur2026image} & \textit{0.948} & \textit{0.935} & \textit{0.917} & \textit{0.699} & \textit{0.738} & \textit{0.793}$^\ast$ \\
\bottomrule
\end{tabular}%
}
\end{table*}

Table~\ref{tab:main} summarizes the performance of the fused initialization and two representative checkpoints from run~B. Continued adaptation substantially improves unaligned NYUv2 $\delta_1$, from 0.278 at initialization to 0.407 at 17.5k updates and 0.497 at 100k updates. This improvement is accompanied by a decline in Cityscapes mIoU, from 0.386 to 0.355 and 0.298, respectively. Between the balanced and depth-oriented checkpoints, referring segmentation also deteriorates: RefCOCOg cIoU decreases from 0.518 to 0.481, and ReasonSeg gIoU from 0.431 to 0.333. Depth gains are themselves dataset-dependent: the final checkpoint achieves the highest unaligned scores among the reported OVB models on NYUv2 and DIODE-Indoor, whereas the initialization remains strongest on ETH3D. Thus, none of the three models dominates across all benchmarks. Qualitative examples are in \cref{sec:s-qual}. 
The contrast between unaligned and median-aligned depth further clarifies the observed limitations. For the final checkpoint, per-image scale correction raises $\delta_1$ from 0.497 to 0.926 on NYUv2, from 0.225 to 0.974 on ETH3D, and from 0.353 to 0.926 on DIODE-Indoor. These large improvements indicate that absolute-scale calibration accounts for a substantial component of the depth error, while much of the relative scene structure is preserved.

\section{Findings}
\label{sec:findings}

\begin{figure*}[t]
\centering
\begin{tikzpicture}
\begin{groupplot}[
  group style={group size=2 by 1, horizontal sep=1.6cm},
  width=0.47\linewidth, height=4.6cm,
  xmin=0, xmax=102, xtick={0,20,40,60,80,100},
  xlabel={training updates ($\times 10^3$)},
  grid=major, grid style={gray!20},
  tick label style={font=\scriptsize}, label style={font=\small},
  title style={font=\small},
  legend style={font=\scriptsize, draw=none, fill=white, fill opacity=0.85, text opacity=1, cells={anchor=west}},
  every axis plot/.append style={line width=1pt, mark size=1.2pt},
]
\nextgroupplot[title={(a) NYUv2 depth: aligned vs.\ unaligned $\delta_1$}, ymin=0, ymax=1.0, ylabel={$\delta_1$}, legend style={at={(0.98,0.70)}, anchor=east}, legend columns=2]
\addplot[armB, mark=*] coordinates {(10,0.924)(20,0.903)(30,0.900)(40,0.886)(50,0.902)(57.5,0.909)(60,0.912)(65,0.923)(67.5,0.924)(70,0.925)(75,0.925)(80,0.924)(85,0.924)(87.5,0.926)(92.5,0.924)(95,0.924)(100,0.925)};
\addlegendentry{B aligned}
\addplot[armA, mark=square*] coordinates {(10,0.913)(30,0.933)(40,0.925)(50,0.900)(60,0.901)(62.5,0.900)(65,0.909)(67.5,0.906)(70,0.904)(72.5,0.905)(75,0.900)(77.5,0.906)(80,0.909)(82.5,0.911)(85,0.911)(87.5,0.911)(90,0.913)(92.5,0.914)(95,0.913)(97.5,0.913)(100,0.912)};
\addlegendentry{A aligned}
\addplot[armB, dashed, mark=o] coordinates {(10,0.111)(20,0.340)(30,0.295)(40,0.266)(50,0.275)(57.5,0.335)(60,0.332)(65,0.283)(67.5,0.279)(70,0.311)(75,0.359)(80,0.384)(85,0.420)(87.5,0.428)(92.5,0.458)(95,0.470)(100,0.485)};
\addlegendentry{B unaligned}
\addplot[armA, dashed, mark=square] coordinates {(10,0.157)(30,0.194)(40,0.251)(50,0.374)(60,0.469)(62.5,0.437)(65,0.410)(67.5,0.387)(70,0.352)(72.5,0.334)(75,0.332)(77.5,0.322)(80,0.311)(82.5,0.312)(85,0.310)(87.5,0.327)(90,0.330)(92.5,0.332)(95,0.332)(97.5,0.328)(100,0.329)};
\addlegendentry{A unaligned}
\addplot[black, densely dotted, line width=0.8pt, mark=none, domain=0:102] {0.948};
\addlegendentry{reference}
\nextgroupplot[title={(b) 2D segmentation}, ymin=0.24, ymax=0.56, ylabel={mIoU / gIoU}, legend style={at={(0.5,0.98)}, anchor=north}, legend columns=2]
\addplot[armB, mark=*] coordinates {(10,0.362)(20,0.353)(30,0.337)(40,0.337)(50,0.328)(57.5,0.324)(60,0.320)(65,0.302)(67.5,0.299)(70,0.295)(75,0.291)(80,0.289)(85,0.293)(87.5,0.291)(92.5,0.294)(95,0.293)(100,0.293)};
\addlegendentry{B Cityscapes}
\addplot[armA, mark=square*] coordinates {(10,0.337)(30,0.309)(40,0.301)(50,0.289)(60,0.276)(62.5,0.274)(65,0.270)(67.5,0.266)(70,0.266)(72.5,0.267)(75,0.266)(77.5,0.265)(80,0.269)(82.5,0.270)(85,0.270)(87.5,0.265)(90,0.264)(92.5,0.264)(95,0.262)(97.5,0.261)(100,0.263)};
\addlegendentry{A Cityscapes}
\addplot[armB, dashed, mark=o] coordinates {(10,0.414)(20,0.440)(30,0.366)(40,0.314)(50,0.314)(57.5,0.335)(60,0.343)(65,0.349)(67.5,0.353)(70,0.355)(75,0.349)(80,0.345)(85,0.338)(87.5,0.338)(92.5,0.333)(95,0.335)(100,0.333)};
\addlegendentry{B ReasonSeg}
\addplot[armA, dashed, mark=square] coordinates {(10,0.301)(30,0.348)(40,0.323)(50,0.417)(60,0.357)(62.5,0.381)(65,0.379)(67.5,0.387)(70,0.398)(72.5,0.397)(75,0.412)(77.5,0.429)(80,0.439)(82.5,0.435)(85,0.434)(87.5,0.428)(90,0.430)(92.5,0.429)(95,0.424)(97.5,0.425)(100,0.427)};
\addlegendentry{A ReasonSeg}
\end{groupplot}
\end{tikzpicture}
\caption{\textbf{Training trajectories of the two 100k-update runs} (A: lr $3\times10^{-5}$, B: lr $1\times10^{-5}$; EMA ``best'' checkpoints, external evaluation). (a)~Median-aligned NYUv2 $\delta_1$ is flat at 0.89--0.93 from the first evaluated checkpoint, while unaligned $\delta_1$ moves between 0.11 and 0.49: training changes the scale, not the geometry. (b)~Cityscapes mIoU declines in both runs while depth improves; ReasonSeg peaks early in B and recovers late in A.}
\label{fig:trajectory}
\end{figure*}
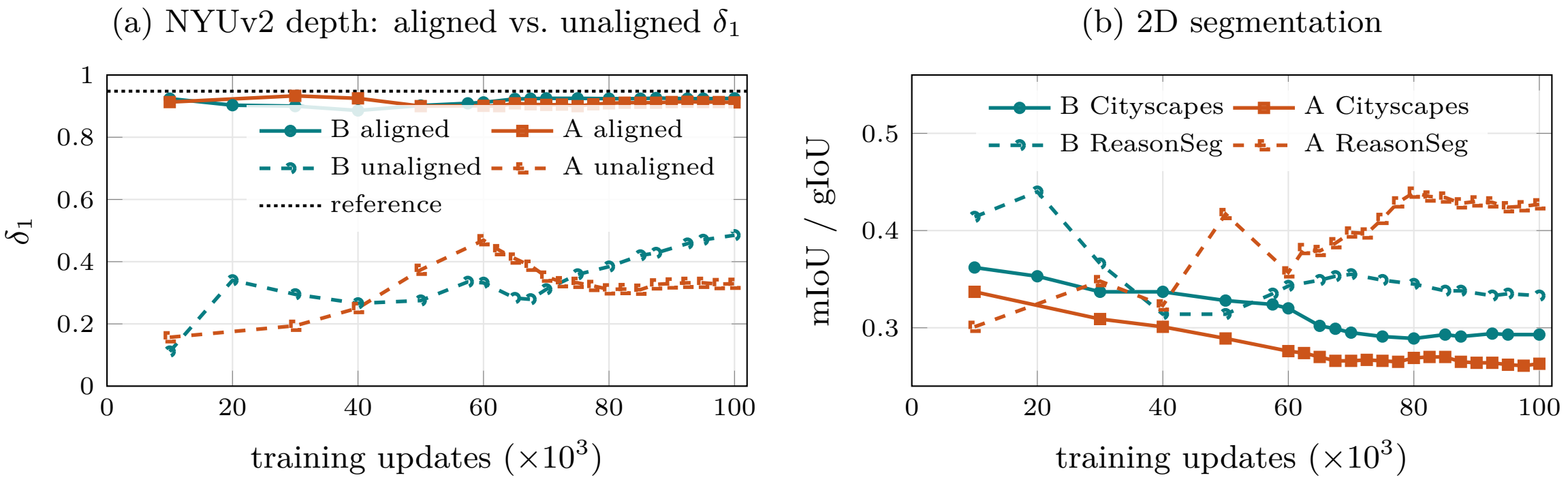

\begin{table*}[!t]
\centering
\small
\caption{\textbf{Controlled 10k-update runs} varying task mixture, colormap, starting point, and adaptation mode (single seed, unaligned $\delta_1$). All runs use an earlier referring prompt and ReasonSeg subset (200 queries), so they are compared only within this table. Top group starts from one earlier LoRA checkpoint (``Parent''); bottom group starts from the untuned Qwen-Image-2.1 backbone. Bold: best per column.}
\label{tab:family}
\setlength{\tabcolsep}{5pt}
\adjustbox{max width=\linewidth}{%
\begin{tabular}{lcccccc}
\toprule
 & \multicolumn{3}{c}{Metric depth, unaligned $\delta_1$} & Semantic & \multicolumn{2}{c}{Referring} \\
\cmidrule(lr){2-4}\cmidrule(lr){5-5}\cmidrule(lr){6-7}
Run & NYUv2 & ETH3D & DIODE-In. & Cityscapes mIoU & RefCOCOg cIoU & ReasonSeg gIoU \\
\midrule
\multicolumn{7}{l}{\textit{From an earlier multi-task checkpoint (Parent)}} \\
Parent (step 0) & 0.134 & 0.122 & 0.109 & 0.342 & 0.498 & \textbf{0.328} \\
C1: depth 75\%, cube-edge only & \textbf{0.348} & 0.233 & \textbf{0.321} & 0.346 & 0.424 & 0.303 \\
T0: all tasks, 10\% replay & 0.131 & 0.237 & 0.216 & 0.335 & 0.438 & 0.287 \\
T1: without normals (22.5\% each + 10\% replay) & 0.149 & 0.192 & 0.180 & 0.337 & 0.467 & 0.271 \\
T2: depth 75\%, five colormaps & 0.179 & \textbf{0.251} & 0.241 & 0.331 & 0.291 & 0.276 \\
T3: segmentation only & 0.050 & 0.032 & 0.143 & 0.264 & 0.519 & 0.304 \\
ML: LoRA, T1 mixture, cube-edge & 0.169 & 0.227 & 0.246 & 0.354 & 0.489 & 0.304 \\
I101-h1: full fine-tuning, T1, cube-edge, lr $5\times10^{-6}$ & 0.222 & 0.193 & 0.247 & 0.362 & 0.241 & 0.175 \\
\midrule
\multicolumn{7}{l}{\textit{From the untuned Qwen-Image-2.1 backbone}} \\
Untuned base (no adapter) & 0.072 & 0.119 & 0.077 & 0.215 & 0.376 & 0.254 \\
I102-i1: fresh LoRA, referring 35\%, cube-edge & 0.099 & 0.195 & 0.115 & \textbf{0.368} & \textbf{0.547} & 0.321 \\
\bottomrule
\end{tabular}%
}
\end{table*}

\subsection{Geometry is learned early; scale is not}
\label{sec:f-scale}

Figure~\ref{fig:trajectory}(a) tracks NYUv2 depth over both runs. Median-aligned $\delta_1$, which corrects one scalar per image using the ground truth, stays between 0.886 and 0.933 at every evaluated checkpoint, already 0.873 at the starting point, and ends at 0.926 in run~B, near the reference's \emph{unaligned} 0.948. Unaligned $\delta_1$ instead moves between 0.111 and 0.497, and in run~A rises to 0.469 at 60k updates and falls back to 0.329. The same holds on the other datasets (all checkpoints are listed in \cref{sec:s-traj}): for the final B adapter, aligned $\delta_1$ is 0.926 / 0.974 / 0.926 on NYUv2 / ETH3D / DIODE-Indoor, against 0.497 / 0.225 / 0.353 unaligned (Table~\ref{tab:main}). The relative structure of the predicted depth is thus close to the reference, and almost the whole remaining error is one number per image: how far away the scene is. The cube-edge code makes this plausible: scale is a global hue shift (Sec.~\ref{sec:method}), which costs little in the latent loss of Eq.~\eqref{eq:flow} but moves every decoded pixel.

\subsection{The model does not commit to a scale}
\label{sec:f-variance}

If scale were a fixed bias of the model, repeated sampling would not change it. We decode $K$ depth maps $\hat{d}^{(k)}$ from $K$ noise seeds, let $m_k$ be the median of member $k$, and combine
\begin{equation}
  \bar{d}_i = \operatorname{med}_k \frac{\hat{d}^{(k)}_i}{m_k}\cdot\exp\Bigl(\operatorname{med}_k \log m_k\Bigr),
  \label{eq:ensemble}
\end{equation}
a pixelwise median of scale-normalized members times the geometric median of their scales, which uses no ground truth. Table~\ref{tab:ensemble} applies it to a depth-heavy earlier run (C1, Table~\ref{tab:family}) without retraining. Unaligned NYUv2 $\delta_1$ rises from 0.348 to 0.436 with five seeds and 0.516 with ten; the gain from five to ten seeds is almost as large as from one to five, so the ensemble has not saturated. Single samples are occasionally catastrophic: on DIODE-Indoor, AbsRel falls from 686.77 to 0.377 because the median discards rare members that decode a whole image to a very large depth. The same image and instruction therefore yield different scales under different noise. How much of the residual gap is a stable per-image bias, which no number of seeds removes, remains open. The ensemble is a diagnostic; at $K$ times the inference cost it is not a proposed method.

\begin{table}[t]
\centering
\small
\caption{\textbf{Seed ensemble} (Eq.~\eqref{eq:ensemble}) on run C1, full validation sets, no retraining. $K{=}1$ is the standard protocol; -- not evaluated.}
\label{tab:ensemble}
\setlength{\tabcolsep}{4pt}
\begin{tabular}{llccc}
\toprule
Dataset & Metric & $K{=}1$ & $K{=}5$ & $K{=}10$ \\
\midrule
NYUv2 & unaligned $\delta_1$ & 0.348 & 0.436 & 0.516 \\
 & aligned $\delta_1$ & 0.845 & -- & 0.939 \\
ETH3D & unaligned $\delta_1$ & 0.233 & -- & 0.354 \\
 & aligned $\delta_1$ & 0.906 & -- & 0.976 \\
DIODE-In. & unaligned $\delta_1$ & 0.321 & -- & 0.392 \\
 & aligned $\delta_1$ & 0.855 & -- & 0.922 \\
 & AbsRel $\downarrow$ & 686.77 & -- & 0.377 \\
\bottomrule
\end{tabular}
\end{table}

\begin{table}[!t]
\centering
\small
\caption{\textbf{Exact weight-space merges} of run-B EMA adapters, $\Delta W=\sum_i w_i\,\Delta W_i$ (rank-$64n$ concatenation, no approximation; P5 weights $.4/.4/.2$, P6 $.2/.55/.25$). Depth is unaligned $\delta_1$; ``Avg.'' is the gap (in points) to the coefficient-weighted average of the parents' scores. All rows use the same protocol; $^\S$P6 is the released BF16 model, scored standalone.}
\label{tab:merge}
\setlength{\tabcolsep}{2.6pt}
\adjustbox{max width=\linewidth}{%
\begin{tabular}{lcccccc}
\toprule
Model ($w$) & NYU & ETH & City & \multicolumn{2}{c}{Reason} & Ref \\
 & & & & gIoU & Avg. & \\
\midrule
B17.5k (EMA) & 0.407 & 0.176 & 0.355 & \textbf{0.431} & -- & \textbf{0.518} \\
B100k (EMA) & \textbf{0.485} & \textbf{0.221} & 0.293 & 0.333 & -- & 0.482 \\
\midrule
P1: 17.5k/100k .75/.25 & 0.407 & 0.193 & 0.346 & 0.400 & $-0.7$ & 0.515 \\
P2: 17.5k/100k .5/.5 & 0.419 & 0.198 & 0.338 & 0.380 & $-0.2$ & 0.506 \\
P3: 5k/100k .6/.4 & 0.431 & 0.214 & 0.350 & 0.371 & $-0.1$ & 0.505 \\
P4: 20k/100k .6/.4 & 0.394 & 0.203 & 0.339 & 0.385 & $-1.2$ & 0.517 \\
P5: 5k/17.5k/100k & 0.406 & 0.194 & \textbf{0.364} & 0.395 & $-0.4$ & 0.513 \\
P6: start/17.5k/100k$^\S$ & 0.450 & 0.200 & 0.363 & 0.399 & $-0.4$ & 0.512 \\
\bottomrule
\end{tabular}%
}
\end{table}

\subsection{Tasks trade off under one adapter}
\label{sec:f-tradeoff}

\paragraph{Along training.}
In run~B (Figure~\ref{fig:trajectory}(b)), Cityscapes mIoU falls from 0.362 at 10k updates to 0.293 at 100k while unaligned NYUv2 $\delta_1$ rises from 0.111 to 0.485; ReasonSeg peaks at 0.440 at 20k and settles near 0.333. The balanced and depth endpoints of Table~\ref{tab:main} are two points of this one trajectory.

\paragraph{Across learning rates.}
The larger learning rate of run~A does not move faster along the same path. At 100k updates A is worse than B on unaligned depth (NYUv2 0.329 vs.\ 0.485) and on Cityscapes (0.263 vs.\ 0.293), but better on ReasonSeg (0.427 vs.\ 0.333). Run~A is also less stable: at five of the six checkpoints between 60k and 72.5k updates its DIODE-Indoor AbsRel lies between 37 and 169, because single images decode to extreme depths.

\paragraph{Across task mixtures.}
Table~\ref{tab:family} varies the mixture, the colormap, the starting point, and the adaptation mode under one protocol (10k updates each). No run improves depth and segmentation together. From the shared parent, the depth-heavy run C1 (75\% depth, cube-edge only) gains 0.214 in NYUv2 $\delta_1$ and loses 0.074 in RefCOCOg; the segmentation-only run T3 lifts RefCOCOg to 0.519 and has the lowest depth scores; and from the untuned backbone, a fresh LoRA weighted toward referring (I102-i1, 35\% referring) sets the best Cityscapes and RefCOCOg scores while leaving NYUv2 at 0.099. At a fixed depth share, a single colormap matters: C1 (cube-edge only) reaches 0.348 on NYUv2 where T2 (five colormaps) reaches 0.179.

\paragraph{Full fine-tuning.}
After merging the parent adapter into the backbone, updating all transformer weights at learning rate $5\times10^{-6}$ on the T1 mixture with cube-edge depth (I101-h1) lowers RefCOCOg cIoU from 0.484 at 2k updates to 0.112 at 4k and ends at 0.241 after 10k, half of the parent's 0.498, while the matched LoRA continuation on the same parent and mixture (ML) keeps RefCOCOg at 0.489. The low-rank constraint is therefore not only a cost saving: at this budget it protects the referring ability of the backbone.

\subsection{Merging moves along the trade-off}
\label{sec:f-merge}

Can weight averaging combine the balanced and depth checkpoints? We merge run-B adapters exactly in weight space, $\Delta W=\sum_i w_i \Delta W_i$ with $\sum_i w_i=1$, by concatenating their low-rank factors (no approximation; verified to $4\times10^{-9}$ against the parents, and a single-parent merge reproduces its parent exactly on a 100-image NYUv2 check). P5 combines the 5k, 17.5k, and 100k adapters with weights 0.4/0.4/0.2, and P6 combines the starting point with the 17.5k and 100k adapters at 0.2/0.55/0.25. Table~\ref{tab:merge} shows three regularities. Every merge improves ETH3D over the balanced checkpoint. Cityscapes is the only metric that gains beyond interpolation, by 0.7--1.7 points over the weighted average of the parents, and the three-way merges reach 0.363--0.364, above both 17.5k (0.355) and 100k (0.293). ReasonSeg, in contrast, falls slightly \emph{below} interpolation in every merge and 3--6 points below the balanced checkpoint. No merge is at least as good as the balanced checkpoint on all benchmarks: merging chooses a point on the trade-off but does not remove it, and the referring ability lost late in training does not come back.

\subsection{Flow loss does not select checkpoints}
\label{sec:f-loss}

Checkpoint selection by validation loss is standard, and our ``best'' checkpoints are selected that way, so the macro-averaged validation flow loss decreases over them by construction. Yet across the EMA checkpoints of each run, the Spearman correlation between loss \emph{reduction} and Cityscapes mIoU is $\rho=-0.88$ (A) and $-0.89$ (B): \emph{lower loss, worse segmentation}. For unaligned NYUv2 $\delta_1$ the sign differs between the runs ($-0.05$ in A, $+0.84$ in B), and likewise for ReasonSeg ($+0.78$ in A, $-0.42$ in B); all correlations are listed in \cref{sec:s-loss}. Across runs, A reaches a lower final loss than B (0.03079 vs.\ 0.03120) while being worse on unaligned depth and Cityscapes. The latent mean squared error is dominated by what is easy to reconstruct, whereas the errors that matter after decoding, the global hue that carries depth scale or a few mislabeled regions, are small in latent space. Decoded metrics on a held-out selection set are needed for model selection.

\section{Discussion and Limitations}
\label{sec:discussion}

\paragraph{Implications.}
The findings translate into three recommendations for generation-as-perception on open models. \emph{(1) Report depth both ways and over seeds.} Unaligned scores conflate geometry with an uncommitted scale; reporting aligned scores and seed spread separates the two. \emph{(2) Select checkpoints by decoded metrics.} The flow loss rewards what is easy to reconstruct in latent space, not the global hue that carries scale or the few regions that decide a mask, so selection needs decoded metrics on a held-out set. \emph{(3) Treat scale and task competition as the targets.} Scale can be supervised where it is decided: reconstruct the clean latent from the predicted velocity, decode it, project onto the cube-edge path, and penalize the per-image log-depth residual. We provide this optional loss in the released code but have not evaluated it, and make no claim about its effect.

\paragraph{Limitations.}
\textbf{Gap to the reference.} Our results are below Vision Banana on every benchmark; we do not reproduce its numbers, and its base model, in-house data, and human evaluation are not public. \textbf{One backbone.} The framework supports other generators through configuration, but all findings are on Qwen-Image-2.1; we do not claim they transfer. \textbf{Selection on evaluation sets.} Checkpoints and merge recipes were inspected on the evaluation benchmarks, so the best reported scores are optimistic by construction: the P5 and P6 weights were fixed after seeing Table~\ref{tab:merge}. Our claims are about the sign of the effect across all six merges, not about any single recipe.

\section{Conclusion}
\label{sec:conclusion}

We built Open Vision Banana, an open framework in which every perception task is a checked contract between an RGB encoder, an instruction, and a decoder, and one adapter on a frozen editing backbone serves all tasks. The open recipe does not match the closed system, but it fails in ways that are now measurable: depth geometry is nearly right while its absolute scale is uncommitted, segmentation competes with depth for a shared adapter, weight merging moves along this trade-off without escaping it, and the training loss cannot tell good checkpoints from bad ones. These failure modes, more than any single benchmark number, mark what an open generator must overcome to become a generalist vision learner; the released framework and evaluation outputs make each of them testable.

\section*{LLM Usage Statement}
We used ChatGPT to assist with code development and organization, the compilation and export of experimental results, and language polishing in selected portions of the manuscript, including grammar correction and improvements to clarity and phrasing.

\bibliographystyle{plainnat}
\bibliography{main}

\clearpage
\beginappendix
This appendix details the evaluation protocol and its comparability with the reference (\cref{sec:s-protocol}), the instruction of every task (\cref{sec:s-prompts}), the instance decoder (\cref{sec:s-instance}), the full training configuration including the fused starting point (\cref{sec:s-training}), the merge construction and full merge results (\cref{sec:s-merge}), the loss--metric correlations (\cref{sec:s-loss}), evaluation noise (\cref{sec:s-noise}), the released artifacts (\cref{sec:s-release}), qualitative results (\cref{sec:s-qual}), and every evaluated checkpoint of the two main runs (\cref{sec:s-traj}). All numbers are taken unchanged from the released evaluation outputs.

\section{Evaluation protocol}
\label{sec:s-protocol}

\paragraph{Inference.}
The source image is resized so that its area matches a $512\times512$ canvas (side a multiple of 32), the target is generated from noise with 50 flow-matching steps and true classifier-free guidance 4.0 against an empty negative prompt, with seed 42. Every raw pipeline output is asserted to be $512\times512$, and for referring tasks every prompt is asserted to match the instruction template of \cref{tab:contracts}. The generated image is decoded by the task decoder and the decoded label or depth map is resized to the native ground-truth resolution (nearest neighbor for labels, so that interpolated colors never create phantom classes; bilinear for depth).

\paragraph{Metric depth.}
The decoder projects each pixel onto the cube-edge path and inverts the power transform (\cref{eq:depth-transform}). Valid pixels are those with finite positive ground truth inside the dataset mask, and with a finite positive prediction after clipping. NYUv2 uses the Eigen crop and a 10\,m cap, ETH3D an 80\,m cap, and DIODE-Indoor no cap. AbsRel, RMSE, and $\delta_1=\Pr[\max(\hat d/d,d/\hat d)<1.25]$ are accumulated over all valid pixels of the dataset (pixel-weighted), not averaged per image. The \emph{aligned} diagnostic rescales each predicted map by $\operatorname{med}(d)/\operatorname{med}(\hat d)$ over its valid pixels before scoring; it uses the ground truth and is never reported as a headline number. Because the prediction-validity filter is applied after decoding, a sample that decodes to extreme depths can dominate DIODE AbsRel, which is why we treat DIODE AbsRel as a stability indicator.

\paragraph{Segmentation.}
Cityscapes predictions are decoded by nearest palette color, label IDs are mapped to the 19 train IDs, and mIoU is computed from one confusion matrix over the 500 validation images using official \texttt{gtFine} annotations. For referring segmentation, the prediction is thresholded to a binary mask; gIoU is the mean IoU over queries, and cIoU is the summed intersection over the summed union. RefCOCOg uses the UMD split (\texttt{refs(umd).p}, 4{,}896 expressions) with native COCO masks. ReasonSeg uses all 344 queries of the validation set with the original query text and the \texttt{legacy\_union} mask rasterization; we also log the official LISA rasterization (\texttt{lisa\_v1}), which is 1--2 points lower.

\begin{table}[!htbp]
\centering
\small
\caption{\textbf{Comparability with the reference.} ``Same'' means the reference defines the benchmark, split, and metric as we use them; ``ours'' marks settings the reference does not disclose.}
\label{tab:s-compare}
\setlength{\tabcolsep}{2.5pt}
\adjustbox{max width=\linewidth}{%
\begin{tabular}{l>{\raggedright\arraybackslash}p{6.4cm}l}
\toprule
Benchmark & Difference from the reference & Status \\
\midrule
NYUv2 / ETH3D / DIODE & crop, cap, pixel aggregation, $512^2$ canvas, single seed & same metric, ours settings \\
Cityscapes & $512^2$ canvas, nearest-color decoding & same metric, ours settings \\
RefCOCOg & instruction template, $512^2$ canvas & same metric, ours settings \\
ReasonSeg & reference rewrites queries with Gemini 2.5 Pro; we use the original query & \textbf{not comparable} \\
\bottomrule
\end{tabular}%
}
\end{table}

\paragraph{Leakage control.}
Training rows are removed when their image identifier appears in any evaluation set, including COCO identifiers of the RefCOCOg UMD validation and test images, which lie inside COCO \texttt{train2014}. NYUv2, ETH3D, and ReasonSeg contribute no training data; DIODE and RefCOCOg contribute only their training splits (Table~\ref{tab:s-compare}).

\section{Task prompts}
\label{sec:s-prompts}

This section lists the instruction of every task verbatim. Training and evaluation use the same templates; placeholders in braces are filled per sample. The number of distinct prompts per task in the training pool is recorded in each checkpoint's metadata; for the run-B checkpoints it is 1 for depth, 145{,}884 for semantic, 724 for instance, 224{,}213 for referring segmentation, and 7{,}251 for editing replay.

\newtcolorbox{promptbox}[1]{enhanced, colback=black!3, colframe=black!35, boxrule=0.5pt, arc=1.2mm, left=2mm, right=2mm, top=1mm, bottom=1mm, fonttitle=\small\bfseries, coltitle=black, colbacktitle=black!10, title={#1}}

\paragraph{Metric depth.} One fixed instruction:
\begin{promptbox}{Depth (training and evaluation)}
\small\ttfamily Generate a metric depth visualization of this image using the cube\_edge colormap.
\end{promptbox}

\paragraph{Semantic segmentation.} Each training row carries its own color legend for the classes in the image, written in one of four formats sampled with probabilities 0.25 (named colors), 0.30 (RGB tuples), 0.25 (sentences), and 0.20 (hex codes); legend colors are remapped with probability 0.5 so that the model must read the legend. With an example legend for three classes:
\begin{promptbox}{Semantic, named colors}
\small\ttfamily Generate a semantic segmentation visualization image, using this color mapping: \{"person": "red", "dog": "green", "background": "blue"\}.
\end{promptbox}
\begin{promptbox}{Semantic, RGB tuples}
\small\ttfamily Generate a semantic segmentation visualization image, using this color mapping: \{"person": <220, 20, 60>, "dog": <0, 0, 142>, "background": <0, 0, 0>\}.
\end{promptbox}
\begin{promptbox}{Semantic, sentences}
\small\ttfamily This image is a per-pixel class labeling of the input. The person are represented by (220, 20, 60). The dog are represented by (0, 0, 142). The background are represented by (0, 0, 0).
\end{promptbox}
\begin{promptbox}{Semantic, hex codes}
\small\ttfamily Generate a semantic segmentation visualization of the input. The person is \#DC143C. The dog is \#00008E. The background is \#000000.
\end{promptbox}
Cityscapes is evaluated with the RGB-tuple format and the full legend of the 19 evaluation classes in the official Cityscapes colors:
\begin{promptbox}{Semantic, Cityscapes evaluation}
\small\ttfamily Generate a semantic segmentation visualization image, using this color mapping: \{"road": <128, 64, 128>, "sidewalk": <244, 35, 232>, "building": <70, 70, 70>, "wall": <102, 102, 156>, "fence": <190, 153, 153>, "pole": <153, 153, 153>, "traffic\_light": <250, 170, 30>, "traffic\_sign": <220, 220, 0>, "vegetation": <107, 142, 35>, "terrain": <152, 251, 152>, "sky": <70, 130, 180>, "person": <220, 20, 60>, "rider": <255, 0, 0>, "car": <0, 0, 142>, "truck": <0, 0, 70>, "bus": <0, 60, 100>, "train": <0, 80, 100>, "motorcycle": <0, 0, 230>, "bicycle": <119, 11, 32>\}.
\end{promptbox}

\paragraph{Instance segmentation.} The instruction names the requested category. Targets on a black background use the first template; targets rendered on another background use one of the other two, which state the background color as a name, an RGB triple, or a hex code. With \texttt{dog} as the category:
\begin{promptbox}{Instance, black background}
\small\ttfamily Generate an instance segmentation visualization of this image. Each dog is colored differently.
\end{promptbox}
\begin{promptbox}{Instance, other backgrounds}
\small\ttfamily This image is a segmentation task derived from the input. The dog instances are each represented by a unique, solid color. Background is RGB(88, 50, 82).\par\medskip
This image shows segmentation masks for the dog from the input image. The background is set to \#10aa05. Each dog is represented by a different color.
\end{promptbox}

\paragraph{Referring segmentation.} One template (\texttt{query\_v2}) for training and evaluation on RefCOCOg and ReasonSeg. The query is inserted unchanged, keeping its words, punctuation, and leading article; ReasonSeg queries are the original ones, without rewriting:
\begin{promptbox}{Referring (training and evaluation)}
\small\ttfamily Identify the object or region referred to by the following query in the image.\\
Query: \{query\}\\
Render its segmentation mask in solid white and all other pixels in black.
\end{promptbox}

\paragraph{Editing replay.} Replay rows use the editing instruction of each source pair unchanged; there is no template.

\section{Instance decoder}
\label{sec:s-instance}

The instance decoder turns a one-color-per-instance RGB prediction on a black background into binary masks in five stages. The parameters are fixed for every result in this paper.
\begin{enumerate}
  \item \textbf{Background initialization.} Pixels within Euclidean RGB distance $\tau=14$ (8-bit scale) of the background color are background.
  \item \textbf{Color grouping.} Foreground pixels are grouped by 16-connected, seed-based region growing: a pixel joins a component when its color lies within $\tau$ of the component's seed color. The grouping is implemented with a union--find.
  \item \textbf{Noise pruning.} Components smaller than $\theta_{\mathrm{size}}=2\times10^{-4}$ of the image area are removed.
  \item \textbf{Boundary-artifact removal.} A component is removed when a $3\times3$ erosion keeps less than $\theta_{\mathrm{erosion}}=0.1$ of its pixels, which removes thin halos along instance boundaries.
  \item \textbf{Spatially constrained merging.} Two components are merged when their mean colors lie within $\tau$ and the bounding box of their union is at most $\gamma=5$ times the sum of their bounding-box areas; merges are applied transitively.
\end{enumerate}
The remaining components are the predicted instances; the released code implements the decoder in the \texttt{SegmentationDecoder} class.

\section{Training configuration}
\label{sec:s-training}

\begin{table}[!htbp]
\centering
\small
\caption{\textbf{Training configuration.} Runs A and B differ only in the peak learning rate. ``K'' is the second parent of the fused starting point.}
\label{tab:s-config}
\setlength{\tabcolsep}{3pt}
\adjustbox{max width=\linewidth}{%
\begin{tabular}{lll}
\toprule
Setting & Runs A / B & Fused parent K \\
\midrule
Adapter & rsLoRA $r{=}64$, $\alpha{=}64$ (scale 8) & rsLoRA $r{=}128$, $\alpha{=}64$ \\
Target modules & \multicolumn{2}{l}{q, k, v, out projections; MLP gate, proj, out} \\
Updates & 100{,}000 & 20{,}000 \\
Peak lr & $3{\times}10^{-5}$ / $1{\times}10^{-5}$ & $3{\times}10^{-5}$ \\
Warmup, schedule & 2{,}000, cosine & 1{,}000, cosine \\
Global batch & 32 ($1{\times}2$ accum.${\times}16$ GPUs) & 32 ($1{\times}4{\times}8$) \\
Optimizer & \multicolumn{2}{l}{AdamW, $\beta{=}(0.9,0.999)$, $\epsilon{=}10^{-8}$, wd 0.01} \\
Clipping & \multicolumn{2}{l}{norm 1.0; skip update if norm $>10$} \\
Timesteps & \multicolumn{2}{l}{logit-normal, shift 1.0, no Min-SNR} \\
Cond.\ dropout & \multicolumn{2}{l}{0.05} \\
EMA & 0.9999, warmup 500 & -- \\
Resolution & \multicolumn{2}{l}{512} \\
Mixture & D .35, S .20, I .20, R .20, E .05 & D, N, S, I, R .18 each, E .10 \\
Seed & \multicolumn{2}{l}{42} \\
\bottomrule
\end{tabular}%
}
\end{table}

Table~\ref{tab:s-config} lists the configuration (D depth, N normals, S semantic, I instance, R referring, E editing replay). The training pool of the main runs contains 407{,}278 depth, 146{,}027 semantic, 233{,}133 instance, 295{,}505 referring, and 16{,}492 editing rows. The referring pool consists of 270{,}195 rows from RefCOCO, RefCOCO+, and RefCOCOg, 11{,}812 from Rea2Seg, and 13{,}498 from the COCO subset of LLM-Seg. Depth uses only the cube-edge colormap; semantic legends use four formats with color remapping probability 0.5.

\paragraph{Fused starting point.}
The frozen transformer of the main runs is $W_0 = W + 0.6\,\Delta W_{\mathrm{C1}} + 0.4\,\Delta W_{\mathrm{K}}$, where $W$ is the shared parent checkpoint, C1 is the depth-heavy run of \cref{tab:family} (10k updates), and K is the rank-128 run of Table~\ref{tab:s-config} (trained on an earlier 212{,}711-row six-task pool: 60{,}852 depth, 8{,}680 normal, 39{,}133 semantic, 38{,}596 instance, 48{,}958 referring, and 16{,}492 editing rows). The effective weights are averaged in FP32 and cast to BF16 once; no parent adapter is loaded afterwards.

\section{Weight-space merges}
\label{sec:s-merge}

\paragraph{Construction.}
All run-B adapters share the starting point $W_0$ and the rsLoRA scale $s=\alpha/\sqrt{r}=8$, so adapter $i$ adds $\Delta W_i = s\,B_iA_i$. For weights $w_i\ge0$ with $\sum_i w_i=1$, the merged model $W_0+\sum_i w_i\Delta W_i$ is represented exactly by one adapter with concatenated factors
\begin{equation}
  A=\begin{bmatrix}A_1\\ \vdots\\ A_n\end{bmatrix},\qquad B=\bigl[\,w_1 s B_1\;\cdots\;w_n s B_n\,\bigr],
\end{equation}
of rank $64n$ and loader scale 1, so $BA=\sum_i w_i s B_iA_i$ with no cross terms. Averaging $A_i$ and $B_i$ separately would introduce cross terms $B_iA_j$ and is not used. For P6, the starting point enters with weight $0.2$ and contributes no delta. Each merge was compared with the weighted sum of its parents on three layers (maximum absolute difference about $4\times10^{-9}$), and a single-parent merge of the 17.5k adapter reproduced that adapter's NYUv2 scores exactly on the first 100 images.

P6 was also exported as a standalone BF16 pipeline---the artifact we release---and scored again in that form. The two evaluations agree within 0.2 points on every metric except DIODE-Indoor AbsRel (0.79 against 0.75, a benchmark whose uncapped decoding lets a few images dominate the pixel average), and their dataset fingerprints match on five of the six benchmarks. We report the BF16 export throughout, since it is what a user of the released model runs.

\begin{table*}[!htbp]
\centering
\scriptsize
\caption{\textbf{Full merge results.} All rows use the same protocol and code revision, except $^\S$P6, scored in BF16 as a standalone model, the form in which it is released, not as an FP32 concatenated adapter; B5k and B17.5k were re-scored together with the merges. Recipes: P1 17.5k/100k .75/.25; P2 .5/.5; P3 5k/100k .6/.4; P4 20k/100k .6/.4; P5 5k/17.5k/100k .4/.4/.2; P6 start/17.5k/100k .2/.55/.25. $^\ddagger$final adapter. ``--'': not evaluated.}
\label{tab:s-merge}
\setlength{\tabcolsep}{3pt}
\begin{tabular}{lcccccccccccc}
\toprule
 & \multicolumn{3}{c}{NYUv2} & \multicolumn{3}{c}{ETH3D} & \multicolumn{2}{c}{DIODE-In.} & Cityscapes & \multicolumn{2}{c}{ReasonSeg} & RefCOCOg \\
\cmidrule(lr){2-4}\cmidrule(lr){5-7}\cmidrule(lr){8-9}\cmidrule(lr){11-12}
Model & AbsRel & $\delta_1$ & aligned & AbsRel & $\delta_1$ & aligned & AbsRel & $\delta_1$ & mIoU & gIoU & cIoU & cIoU \\
\midrule
B5k & 0.347 & 0.411 & 0.879 & 0.514 & 0.205 & 0.930 & 0.51 & 0.313 & 0.373 & 0.399 & 0.292 & -- \\
B17.5k & 0.294 & 0.407 & 0.921 & 0.515 & 0.176 & 0.932 & 0.51 & 0.277 & 0.355 & 0.431 & 0.328 & 0.518 \\
B20k & 0.362 & 0.340 & 0.903 & 0.600 & 0.166 & 0.910 & 1.07 & 0.229 & 0.353 & 0.440 & -- & -- \\
B100k & 0.267 & 0.485 & 0.925 & 0.418 & 0.221 & 0.974 & 0.47 & 0.345 & 0.293 & 0.333 & -- & 0.482 \\
B100k$^\ddagger$ & 0.263 & 0.497 & 0.926 & 0.412 & 0.225 & 0.974 & 0.42 & 0.353 & 0.298 & 0.333 & -- & 0.481 \\
\midrule
P1 & 0.295 & 0.407 & 0.919 & 0.488 & 0.193 & 0.950 & 0.56 & 0.283 & 0.346 & 0.400 & 0.289 & 0.515 \\
P2 & 0.288 & 0.419 & 0.921 & 0.473 & 0.198 & 0.959 & 2.69 & 0.293 & 0.338 & 0.380 & 0.275 & 0.506 \\
P3 & 0.311 & 0.431 & 0.900 & 0.478 & 0.214 & 0.949 & 1.13 & 0.282 & 0.350 & 0.371 & 0.274 & 0.505 \\
P4 & 0.320 & 0.394 & 0.912 & 0.506 & 0.203 & 0.951 & 2.88 & 0.284 & 0.339 & 0.385 & 0.288 & 0.517 \\
P5 & 0.306 & 0.406 & 0.903 & 0.504 & 0.194 & 0.940 & 0.50 & 0.262 & 0.364 & 0.395 & 0.288 & 0.513 \\
P6$^\S$ & 0.324 & 0.450 & 0.901 & 0.484 & 0.200 & 0.949 & 0.79 & 0.300 & 0.363 & 0.399 & 0.301 & 0.512 \\
\bottomrule
\end{tabular}
\end{table*}

\begin{table}[!htbp]
\centering
\small
\caption{\textbf{Merges relative to interpolation.} Measured score minus the coefficient-weighted average of the parents' scores, in points. ``--'': a parent lacks a same-protocol score.}
\label{tab:s-interp}
\setlength{\tabcolsep}{4pt}
\begin{tabular}{lccccc}
\toprule
 & NYU $\delta_1$ & ETH $\delta_1$ & City & Reason & Ref \\
\midrule
P1 & $-1.96$ & $+0.49$ & $+0.66$ & $-0.68$ & $+0.61$ \\
P2 & $-2.73$ & $-0.13$ & $+1.41$ & $-0.21$ & $+0.63$ \\
P3 & $-0.98$ & $+0.22$ & $+0.90$ & $-0.14$ & -- \\
P4 & $-0.43$ & $+1.43$ & $+0.96$ & $-1.20$ & -- \\
P5 & $-1.86$ & $-0.33$ & $+1.41$ & $-0.38$ & -- \\
P6$^\S$ & $+4.96$ & $-1.03$ & $+1.72$ & $-0.43$ & -- \\
\bottomrule
\end{tabular}
\end{table}

Table~\ref{tab:s-merge} gives all metrics and Table~\ref{tab:s-interp} the gap to interpolation. Cityscapes is above interpolation for every merge, ReasonSeg below for every merge, and unaligned NYUv2 $\delta_1$ below for all merges except P6. P2, P3, and P4 show unstable DIODE AbsRel (1.1--2.9), while P1 and P5 stay at 0.50--0.56, close to the 17.5k checkpoint (0.51), and P6 is 0.79.

\section{Validation loss and decoded metrics}
\label{sec:s-loss}

\begin{table}[!htbp]
\centering
\small
\caption{\textbf{Spearman correlation between loss reduction and decoded metrics} over the EMA ``best'' checkpoints of each run (A: 21, B: 17). Positive: lower loss comes with a better score.}
\label{tab:s-loss}
\setlength{\tabcolsep}{3pt}
\begin{tabular}{lcccccc}
\toprule
Run & NYU $\delta_1$ & NYU aligned & ETH $\delta_1$ & DIODE $\delta_1$ & City & Reason \\
\midrule
A & $-0.05$ & $+0.20$ & $+0.05$ & $-0.36$ & $-0.88$ & $+0.78$ \\
B & $+0.84$ & $+0.62$ & $+0.80$ & $+0.91$ & $-0.89$ & $-0.42$ \\
\bottomrule
\end{tabular}
\end{table}

Table~\ref{tab:s-loss} extends \cref{sec:f-loss}. Only Cityscapes has a consistent sign, and it is negative in both runs. Across runs, A ends at a lower loss than B ($3.079$ vs.\ $3.120\times10^{-2}$) while being worse on unaligned depth and Cityscapes.

\section{Evaluation noise}
\label{sec:s-noise}

\begin{table}[!htbp]
\centering
\small
\caption{\textbf{Repeated evaluation of identical weights.} Top: the 17.5k adapter scored twice by the same suite at two code revisions. Bottom: ReasonSeg gIoU of eight identical adapters (matching SHA-256) scored by two independent evaluation setups.}
\label{tab:s-noise}
\setlength{\tabcolsep}{4pt}
\begin{tabular}{lccc}
\toprule
B17.5k & first & second & $\Delta$ (pt) \\
\midrule
NYU AbsRel & 0.2947 & 0.2944 & $-0.0003$ \\
NYU $\delta_1$ & 0.4069 & 0.4067 & $-0.02$ \\
ETH3D $\delta_1$ & 0.1767 & 0.1764 & $-0.03$ \\
Cityscapes mIoU & 0.3567 & 0.3551 & $-0.16$ \\
ReasonSeg gIoU & 0.4335 & 0.4310 & $-0.25$ \\
RefCOCOg cIoU & 0.5182 & 0.5182 & $\phantom{-}0.00$ \\
\midrule
\multicolumn{3}{l}{ReasonSeg, 8 identical adapters, two setups} & $-0.22$ to $+0.38$ \\
\bottomrule
\end{tabular}
\end{table}

Differences below about 0.4 points between two evaluations of the same weights arise from software and labeling details alone (Table~\ref{tab:s-noise}). We therefore do not interpret differences of this size between models, and the findings of this paper rest on effects of several points or on consistent trends across many checkpoints.

\section{Released artifacts}
\label{sec:s-release}

All artifacts are public; model and data repositories are on the Hugging Face Hub and keep the licenses of their upstream data and base models.
\begin{itemize}
  \item \textbf{Code}: training and evaluation code with the task contracts and pre-training checks, and all configurations, at \url{https://github.com/isLinXu/open-vision-banana}; project page \url{https://islinxu.github.io/open-vision-banana/}.
  \item \textbf{Fused starting point} $W_0$: \href{https://huggingface.co/Melo111/Fusion-C1-KQ0r128-60-40}{\nolinkurl{Melo111/Fusion-C1-KQ0r128-60-40}}.
  \item \textbf{Adapters of runs A and B}, including every evaluated checkpoint: \href{https://huggingface.co/khazic/ovb-i108-a-lr3e-5-r64-2x8-s42_20261109r3}{\nolinkurl{khazic/ovb-i108-a-lr3e-5-r64-2x8-s42_20261109r3}} and \href{https://huggingface.co/khazic/ovb-i108-b-lr1e-5-r64-2x8-s42_20261109r3}{\nolinkurl{khazic/ovb-i108-b-lr1e-5-r64-2x8-s42_20261109r3}}.
  \item \textbf{Merges}: the merge adapters with their evaluation outputs and SHA-256 checksums, \href{https://huggingface.co/khazic/ovb-i127-merge-eval}{\nolinkurl{khazic/ovb-i127-merge-eval}}; P6 as a standalone BF16 pipeline that needs no adapter, \href{https://huggingface.co/Melo111/Fusion-I108-P6}{\nolinkurl{Melo111/Fusion-I108-P6}}.
  \item \textbf{Evaluation outputs of the main runs}: one JSON file per checkpoint and benchmark, with dataset fingerprints and sample counts, \href{https://huggingface.co/khazic/ovb-i120-external-eval}{\nolinkurl{khazic/ovb-i120-external-eval}}.
  \item \textbf{Training data}: RefCOCO-family referring targets, \href{https://huggingface.co/datasets/Melo111/open-vision-banana-res}{\nolinkurl{Melo111/open-vision-banana-res}}; depth and reasoning supplements, \href{https://huggingface.co/datasets/Melo111/open-vision-banana-depth}{\nolinkurl{Melo111/open-vision-banana-depth}}.
\end{itemize}

\section{Qualitative results}
\label{sec:s-qual}

\begin{figure}[!htbp]
\centering
\setlength{\tabcolsep}{2pt}
\begin{tabular}{@{}lccc@{}}
 & \small Input & \small Prediction & \small Ground truth \\
\parbox[c]{2.6cm}{\raggedright\small\textbf{Metric depth}\\\footnotesize NYUv2} & \raisebox{-0.5\height}{\includegraphics[width=3.6cm]{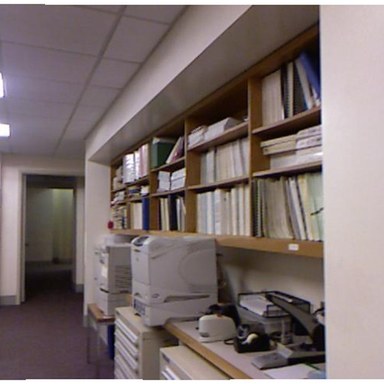}} & \raisebox{-0.5\height}{\includegraphics[width=3.6cm]{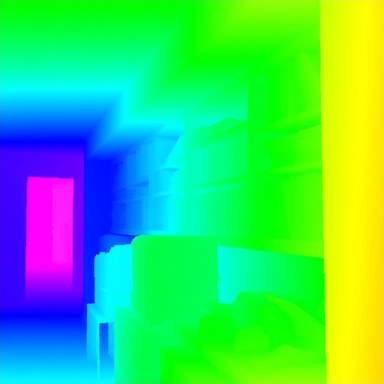}} & \raisebox{-0.5\height}{\includegraphics[width=3.6cm]{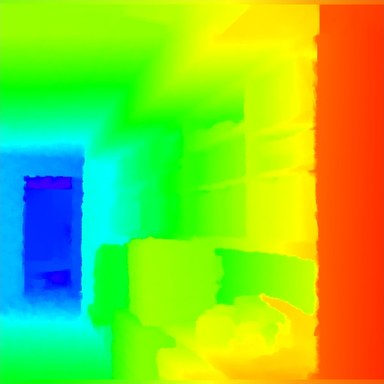}} \\[2pt]
\parbox[c]{2.6cm}{\raggedright\small\textbf{Metric depth}\\\footnotesize DIODE-Indoor} & \raisebox{-0.5\height}{\includegraphics[width=3.6cm]{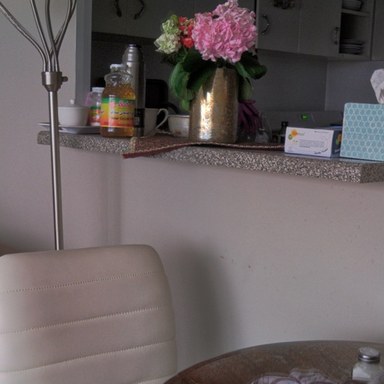}} & \raisebox{-0.5\height}{\includegraphics[width=3.6cm]{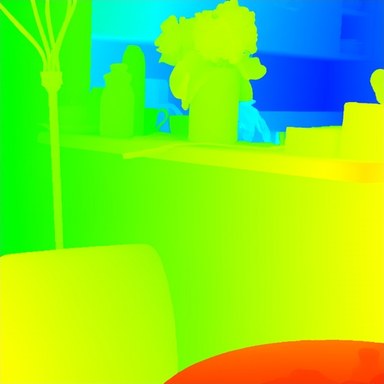}} & \raisebox{-0.5\height}{\includegraphics[width=3.6cm]{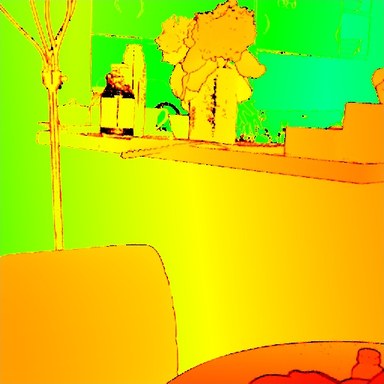}} \\[2pt]
\parbox[c]{2.6cm}{\raggedright\small\textbf{Semantic}\\\footnotesize Cityscapes} & \raisebox{-0.5\height}{\includegraphics[width=3.6cm]{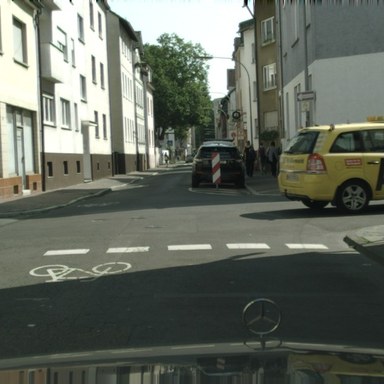}} & \raisebox{-0.5\height}{\includegraphics[width=3.6cm]{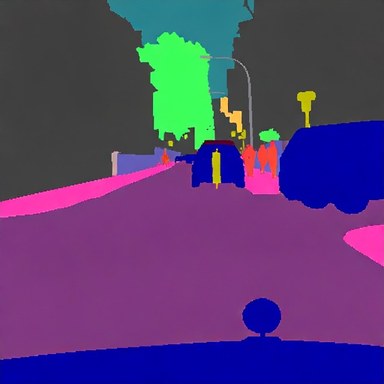}} & \raisebox{-0.5\height}{\includegraphics[width=3.6cm]{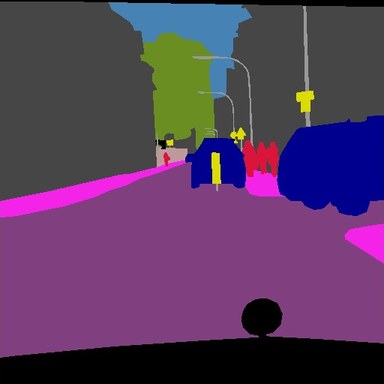}} \\[2pt]
\end{tabular}
\caption{\textbf{Qualitative results: metric depth and semantic segmentation} of Qwen-Image-2.1 with run~B at 20k updates (EMA adapter, merged) on benchmark validation images. Depth is the RGB code before decoding.}
\label{fig:s-qual}
\end{figure}

\begin{figure}[!htbp]
\centering
\setlength{\tabcolsep}{2pt}
\begin{tabular}{@{}lccc@{}}
 & \small Input & \small Prediction & \small Ground truth \\
\parbox[c]{2.6cm}{\raggedright\small\textbf{Instance}\\\footnotesize COCO val2017} & \raisebox{-0.5\height}{\includegraphics[width=3.6cm]{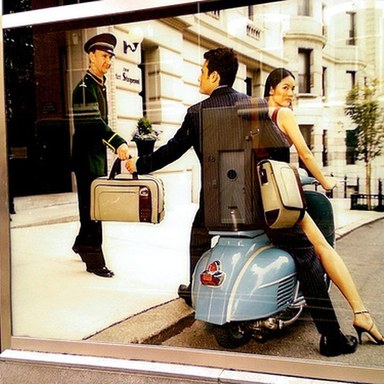}} & \raisebox{-0.5\height}{\includegraphics[width=3.6cm]{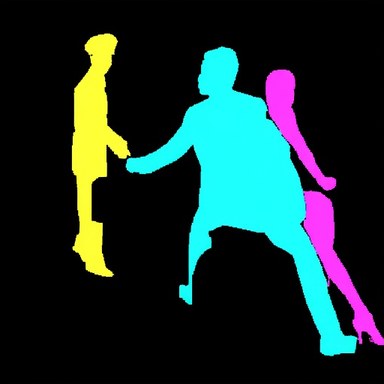}} & \raisebox{-0.5\height}{\includegraphics[width=3.6cm]{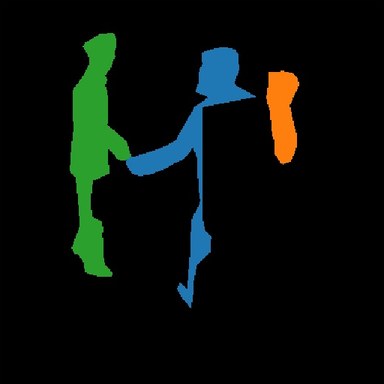}} \\[2pt]
\parbox[c]{2.6cm}{\raggedright\small\textbf{Instance}\\\footnotesize LVIS v1 val} & \raisebox{-0.5\height}{\includegraphics[width=3.6cm]{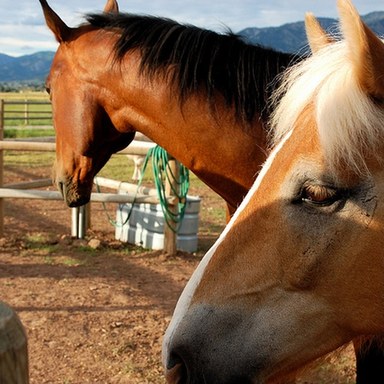}} & \raisebox{-0.5\height}{\includegraphics[width=3.6cm]{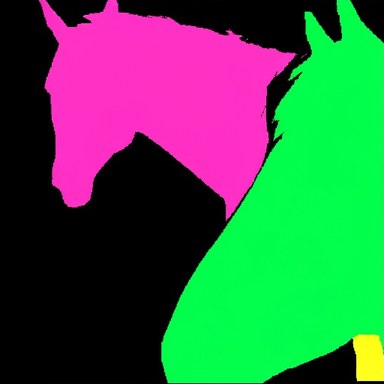}} & \raisebox{-0.5\height}{\includegraphics[width=3.6cm]{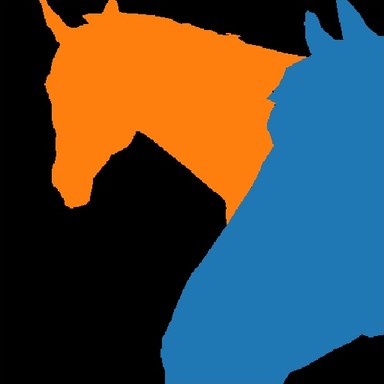}} \\[2pt]
\parbox[c]{2.6cm}{\raggedright\small\textbf{Referring}\\\footnotesize ReasonSeg val} & \raisebox{-0.5\height}{\includegraphics[width=3.6cm]{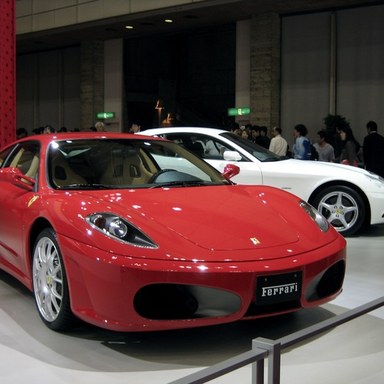}} & \raisebox{-0.5\height}{\includegraphics[width=3.6cm]{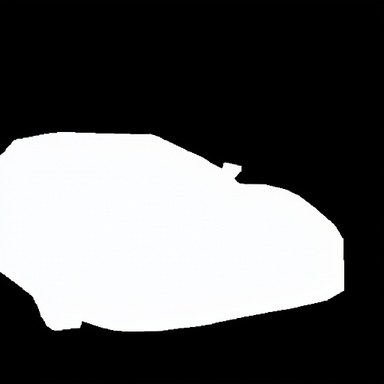}} & \raisebox{-0.5\height}{\includegraphics[width=3.6cm]{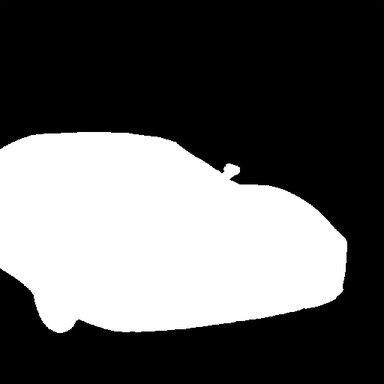}} \\[2pt]
\parbox[c]{2.6cm}{\raggedright\small\textbf{Referring}\\\footnotesize RefCOCOg val} & \raisebox{-0.5\height}{\includegraphics[width=3.6cm]{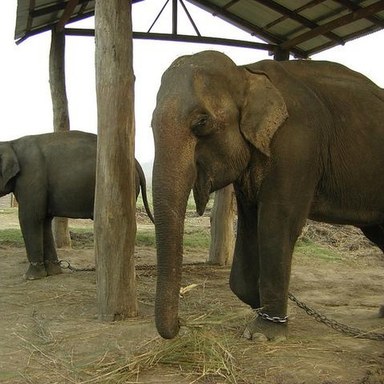}} & \raisebox{-0.5\height}{\includegraphics[width=3.6cm]{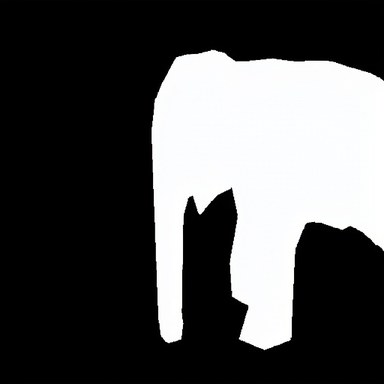}} & \raisebox{-0.5\height}{\includegraphics[width=3.6cm]{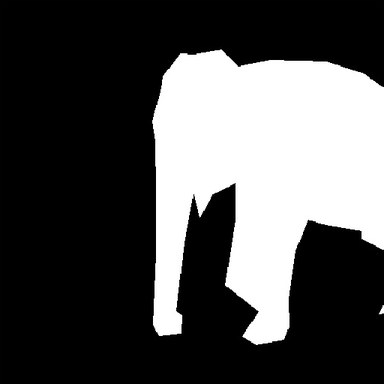}} \\[2pt]
\end{tabular}
\caption{\textbf{Qualitative results: instance and referring segmentation}, same model as \cref{fig:s-qual}. Instance colors are arbitrary per instance and need not match the ground truth.}
\label{fig:s-qual2}
\end{figure}

\begin{figure}[!htbp]
\centering
\begin{minipage}[t]{0.32\linewidth}\centering\includegraphics[width=0.49\linewidth]{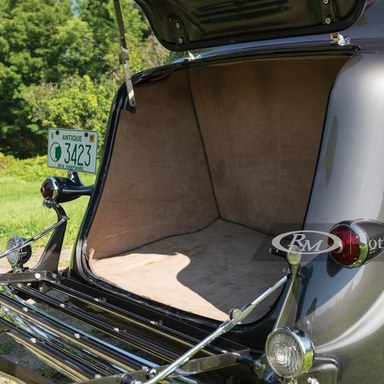}\hfill\includegraphics[width=0.49\linewidth]{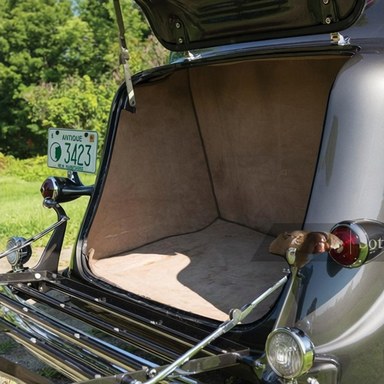}\\[2pt]{\footnotesize ImgEdit-Bench: \textit{``Add a small brown dog sitting inside the open trunk of the antique car.''}}\end{minipage} \hfill \begin{minipage}[t]{0.32\linewidth}\centering\includegraphics[width=0.49\linewidth]{figures/gallery/edit_coco_watercolor_input.jpg}\hfill\includegraphics[width=0.49\linewidth]{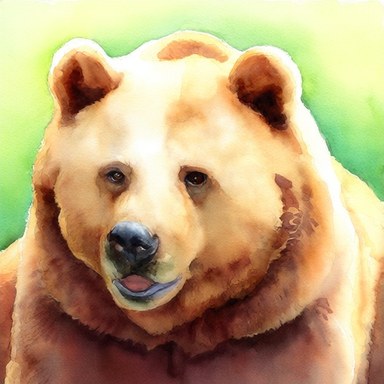}\\[2pt]{\footnotesize COCO val2017: \textit{``Turn this photograph into a watercolor painting while preserving the composition and recognizable objects.''}}\end{minipage} \hfill \begin{minipage}[t]{0.32\linewidth}\centering\includegraphics[width=0.49\linewidth]{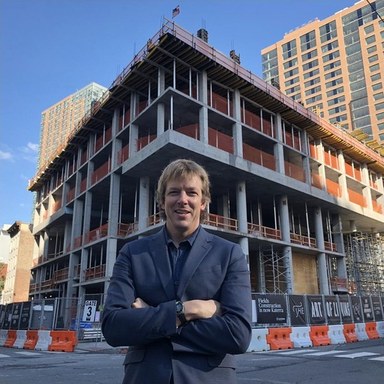}\hfill\includegraphics[width=0.49\linewidth]{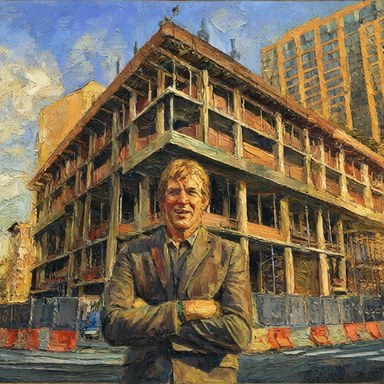}\\[2pt]{\footnotesize ImgEdit-Bench: \textit{``Transfer the image into a classic impasto oil-painting style.''}}\end{minipage}
\caption{\textbf{Editing with the shared adapter.} Each pair shows the input (left) and the output (right) for the given instruction. Qualitative only; generation and editing quality are not measured in this paper.}
\label{fig:s-edit}
\end{figure}

\Cref{fig:s-qual,fig:s-qual2} show predictions of Qwen-Image-2.1 with run~B at 20k updates (EMA adapter merged into a standalone BF16 pipeline, \href{https://huggingface.co/gatilin/ovb-b-20000-best-full}{\nolinkurl{gatilin/ovb-b-20000-best-full}}) on validation and test images of the evaluation benchmarks; the same samples are shown in the project repository. Depth is shown as the RGB code the model writes, before it is decoded to meters. Instance colors are assigned per instance and need not match the colors of the ground truth, which the decoder does not use. The samples were picked for illustration and are not a random draw; the benchmark scores of this checkpoint (row B20k) are in \cref{tab:s-merge}, and those of all checkpoints in \cref{sec:s-traj}. \Cref{fig:s-edit} shows the same adapter following ordinary editing instructions; it is a qualitative check only, since we do not measure generation or editing quality (\cref{sec:discussion}).

\clearpage
\section{All checkpoints of the main runs}
\label{sec:s-traj}

\begin{table*}[!htbp]
\centering
\scriptsize
\caption{\textbf{Checkpoints of runs A ($3{\times}10^{-5}$) and B ($1{\times}10^{-5}$) analysed in this paper.} Depth: unaligned $\delta_1$, median-aligned $\delta_1$ (diagnostic), and unaligned AbsRel. ReasonSeg: 344 queries, \texttt{legacy\_union}. Val.\ loss: macro-averaged validation flow loss $\times10^3$.}
\label{tab:s-traj}
\setlength{\tabcolsep}{3pt}
\begin{tabular}{llccccccccccc}
\toprule
 & & \multicolumn{3}{c}{NYUv2} & \multicolumn{2}{c}{ETH3D} & \multicolumn{3}{c}{DIODE-Indoor} & Cityscapes & ReasonSeg & Val. \\
\cmidrule(lr){3-5}\cmidrule(lr){6-7}\cmidrule(lr){8-10}
Run & k-upd. & $\delta_1$ & aligned & AbsRel & $\delta_1$ & aligned & $\delta_1$ & aligned & AbsRel & mIoU & gIoU & loss \\
\midrule
start & 0 & 0.278 & 0.873 & 0.72 & 0.292 & 0.924 & 0.300 & 0.844 & 3.79 & 0.386 & 0.417 & -- \\
\midrule
A & 10 & 0.157 & 0.913 & 0.39 & 0.075 & 0.931 & 0.169 & 0.899 & 0.47 & 0.337 & 0.301 & 31.752 \\
A & 30 & 0.194 & 0.933 & 0.37 & 0.104 & 0.938 & 0.195 & 0.911 & 0.49 & 0.309 & 0.348 & 31.414 \\
A & 40 & 0.251 & 0.925 & 0.34 & 0.112 & 0.949 & 0.155 & 0.928 & 0.48 & 0.301 & 0.323 & 31.254 \\
A & 50 & 0.374 & 0.900 & 0.36 & 0.195 & 0.940 & 0.253 & 0.901 & 0.72 & 0.289 & 0.417 & 31.175 \\
A & 60 & 0.469 & 0.901 & 0.40 & 0.258 & 0.958 & 0.316 & 0.890 & 168.8 & 0.276 & 0.357 & 31.005 \\
A & 62.5 & 0.437 & 0.900 & 0.39 & 0.255 & 0.962 & 0.300 & 0.889 & 118.6 & 0.274 & 0.381 & 30.979 \\
A & 65 & 0.410 & 0.909 & 0.36 & 0.211 & 0.965 & 0.284 & 0.890 & 143.4 & 0.270 & 0.379 & 30.944 \\
A & 67.5 & 0.387 & 0.906 & 0.37 & 0.200 & 0.964 & 0.270 & 0.892 & 37.2 & 0.266 & 0.387 & 30.941 \\
A & 70 & 0.352 & 0.904 & 0.38 & 0.192 & 0.960 & 0.242 & 0.893 & 1.61 & 0.266 & 0.398 & 30.912 \\
A & 72.5 & 0.334 & 0.905 & 0.39 & 0.177 & 0.958 & 0.227 & 0.890 & 101.2 & 0.267 & 0.397 & 30.887 \\
A & 75 & 0.332 & 0.900 & 0.40 & 0.178 & 0.957 & 0.213 & 0.895 & 2.02 & 0.266 & 0.412 & 30.865 \\
A & 77.5 & 0.322 & 0.906 & 0.39 & 0.177 & 0.956 & 0.207 & 0.899 & 1.43 & 0.265 & 0.429 & 30.855 \\
A & 80 & 0.311 & 0.909 & 0.38 & 0.182 & 0.956 & 0.196 & 0.906 & 0.84 & 0.269 & 0.439 & 30.841 \\
A & 82.5 & 0.312 & 0.911 & 0.37 & 0.187 & 0.959 & 0.195 & 0.907 & 0.90 & 0.270 & 0.435 & 30.839 \\
A & 85 & 0.310 & 0.911 & 0.36 & 0.183 & 0.960 & 0.189 & 0.907 & 1.32 & 0.270 & 0.434 & 30.822 \\
A & 87.5 & 0.327 & 0.911 & 0.35 & 0.186 & 0.966 & 0.192 & 0.913 & 0.91 & 0.265 & 0.428 & 30.812 \\
A & 90 & 0.330 & 0.913 & 0.34 & 0.183 & 0.966 & 0.196 & 0.912 & 0.96 & 0.264 & 0.430 & 30.811 \\
A & 92.5 & 0.332 & 0.914 & 0.33 & 0.183 & 0.967 & 0.194 & 0.911 & 2.30 & 0.264 & 0.429 & 30.804 \\
A & 95 & 0.332 & 0.913 & 0.33 & 0.183 & 0.961 & 0.191 & 0.912 & 1.01 & 0.262 & 0.424 & 30.796 \\
A & 97.5 & 0.328 & 0.913 & 0.33 & 0.184 & 0.963 & 0.185 & 0.912 & 0.83 & 0.261 & 0.425 & 30.795 \\
A & 100 & 0.329 & 0.912 & 0.33 & 0.185 & 0.962 & 0.183 & 0.913 & 1.01 & 0.263 & 0.427 & 30.793 \\
A & 100$^\ddagger$ & 0.326 & 0.912 & 0.34 & 0.184 & 0.962 & 0.173 & 0.915 & 0.78 & 0.261 & 0.423 & -- \\
\midrule
B & 10 & 0.111 & 0.924 & 0.44 & 0.056 & 0.908 & 0.093 & 0.901 & 0.62 & 0.362 & 0.414 & 31.781 \\
B & 20 & 0.340 & 0.903 & 0.36 & 0.166 & 0.910 & 0.229 & 0.888 & 1.07 & 0.353 & 0.440 & 31.620 \\
B & 30 & 0.295 & 0.900 & 0.41 & 0.180 & 0.942 & 0.183 & 0.889 & 58.3 & 0.337 & 0.366 & 31.501 \\
B & 40 & 0.266 & 0.886 & 0.43 & 0.154 & 0.936 & 0.167 & 0.901 & 16.8 & 0.337 & 0.314 & 31.384 \\
B & 50 & 0.275 & 0.902 & 0.38 & 0.174 & 0.940 & 0.192 & 0.911 & 0.88 & 0.328 & 0.314 & 31.337 \\
B & 57.5 & 0.335 & 0.909 & 0.35 & 0.192 & 0.955 & 0.255 & 0.920 & 1.84 & 0.324 & 0.335 & 31.313 \\
B & 60 & 0.332 & 0.912 & 0.34 & 0.190 & 0.954 & 0.250 & 0.920 & 2.74 & 0.320 & 0.343 & 31.310 \\
B & 65 & 0.283 & 0.923 & 0.34 & 0.170 & 0.952 & 0.222 & 0.924 & 2.07 & 0.302 & 0.349 & 31.297 \\
B & 67.5 & 0.279 & 0.924 & 0.34 & 0.164 & 0.948 & 0.227 & 0.924 & 2.00 & 0.299 & 0.353 & 31.289 \\
B & 70 & 0.311 & 0.925 & 0.32 & 0.169 & 0.955 & 0.242 & 0.925 & 1.90 & 0.295 & 0.355 & 31.287 \\
B & 75 & 0.359 & 0.925 & 0.31 & 0.182 & 0.961 & 0.264 & 0.924 & 0.51 & 0.291 & 0.349 & 31.278 \\
B & 80 & 0.384 & 0.924 & 0.30 & 0.187 & 0.962 & 0.283 & 0.924 & 0.46 & 0.289 & 0.345 & 31.254 \\
B & 85 & 0.420 & 0.924 & 0.28 & 0.197 & 0.967 & 0.299 & 0.923 & 0.46 & 0.293 & 0.338 & 31.235 \\
B & 87.5 & 0.428 & 0.926 & 0.28 & 0.205 & 0.970 & 0.304 & 0.923 & 0.65 & 0.291 & 0.338 & 31.219 \\
B & 92.5 & 0.458 & 0.924 & 0.27 & 0.214 & 0.972 & 0.325 & 0.925 & 0.61 & 0.294 & 0.333 & 31.208 \\
B & 95 & 0.470 & 0.924 & 0.27 & 0.216 & 0.974 & 0.334 & 0.924 & 0.60 & 0.293 & 0.335 & 31.207 \\
B & 100 & 0.485 & 0.925 & 0.27 & 0.221 & 0.974 & 0.345 & 0.926 & 0.47 & 0.293 & 0.333 & 31.202 \\
B & 100$^\ddagger$ & 0.497 & 0.926 & 0.26 & 0.225 & 0.974 & 0.353 & 0.926 & 0.42 & 0.298 & 0.333 & -- \\
\bottomrule
\end{tabular}
\end{table*}

Beyond the rows listed here, the released evaluation suite (\cref{sec:s-release}) contains further checkpoints of both runs that were still queued when this table was assembled. The additional checkpoints we inspected continue the trends above rather than reversing them: at 2.5k updates run~A reaches Cityscapes mIoU 0.359 and ReasonSeg gIoU 0.357, against 0.386 and 0.417 before adaptation, while at 12.5k updates run~B reaches Cityscapes mIoU 0.376 and NYUv2 $\delta_1$ 0.189. These points are not used in the analysis of \cref{sec:findings}; they are released so that the trajectory of both runs can be checked in full.

\end{document}